\documentclass[journal]{IEEEtran}
\usepackage{amsmath,amsfonts}
\usepackage{amsmath}
\usepackage{tikz}
\usepackage{algorithmic}
\usepackage{algorithm}
\usepackage{array}
\usepackage[caption=false,font=normalsize,labelfont=sf,textfont=sf]{subfig}
\usepackage{textcomp}
\usepackage{stfloats}
\usepackage{url}
\usepackage{verbatim}
\usepackage{booktabs}
\usepackage{makecell}
\usepackage{graphicx}
\usepackage{cite}
\usepackage{multirow}
\usepackage{xcolor}
\usepackage[colorlinks,
            linkcolor=blue,
            anchorcolor=blue,
            citecolor=blue]{hyperref}
\usepackage{orcidlink}

\begin{document}
\title{A Semantic-Aware and Multi-Guided Network\\ for Infrared-Visible Image Fusion}

\author{Xiaoli Zhang,~\IEEEmembership{Member,~IEEE}, Liying Wang, Libo Zhao, Xiongfei Li, Siwei Ma,~\IEEEmembership{Fellow,~IEEE}

\thanks{Corresponding author: Siwei Ma.}
\thanks{Xiaoli Zhang, Liying Wang, Libo Zhao and Xiongfei Li are with the Key Laboratory of Symbolic Computation and Knowledge Engineering of Ministry of Education, Jilin University, Changchun 130012, China. }

\thanks{ Siwei Ma is with the Institute of Digital Media, School of Electronic Engineering and Computer Science, Peking University, Beijing 100871, China. }
}

\markboth{Journal of \LaTeX\ Class Files,~Vol.~14, No.~8, August~2021}%
{Shell \MakeLowercase{\textit{et al.}}: A Sample Article Using IEEEtran.cls for IEEE Journals}


\maketitle

\begin{abstract}
Multi-modality image fusion aims at fusing modality-specific (complementarity) and modality-shared (correlation) information from multiple source images. To tackle the problem of the neglect of inter-feature relationships, high-frequency information loss, and the limited attention to downstream tasks, this paper focuses on 
how to model correlation-driven decomposing features and reason high-level graph representation by
efficiently extracting complementary information and aggregating multi-guided features. We propose a three-branch encoder-decoder architecture along with corresponding fusion layers as the fusion strategy. 
Firstly, shallow features from individual modalities are extracted by a depthwise convolution layer combined with the transformer block. In the three parallel branches of the encoder, Cross Attention and Invertible Block (CAI) extracts local features and preserves high-frequency texture details. Base Feature Extraction Module (BFE) captures long-range dependencies and enhances modality-shared information. Graph Reasoning Module (GR) is introduced to reason high-level cross-modality relations and simultaneously extract low-level detail features as CAI's modality-specific complementary information. Experiments demonstrate the competitive results compared with state-of-the-art methods in visible/infrared image fusion and medical image fusion tasks. Moreover, the proposed algorithm surpasses the state-of-the-art methods in terms of subsequent tasks, averagely scoring 8.27\% mAP@0.5 higher in object detection and 5.85\% mIoU higher in semantic segmentation. The code is avaliable at \url{https://github.com/Abraham-Einstein/SMFNet/}.

\end{abstract}


\begin{figure}[!htbp]
    \centering
    \includegraphics[width=\linewidth]{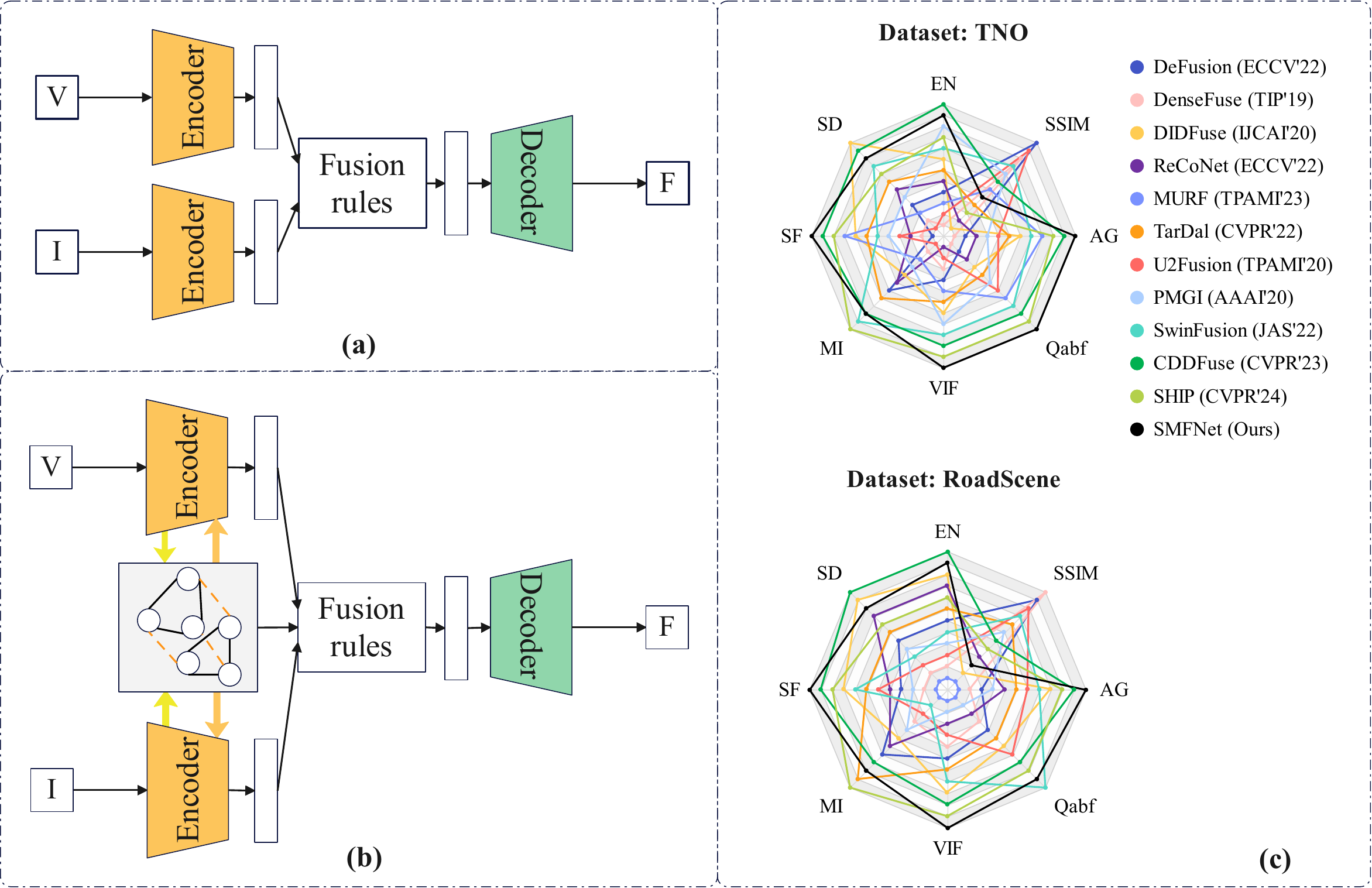}
    \caption{(a) Pipeline of typical Auto-Encoder-based methods, consisting of encoders, a decoder and auxiliary fusion strategies; (b) Our proposed semantic-aware and multi-guided network; (c) We show the qualitative results of fused images on TNO and RoadScene datasets compared with eleven state-of-the-art methods via the radar plots.}
    \label{1}
\end{figure}

\section{Introduction}
\IEEEPARstart{M}{ulti-modal} image fusion has attracted widespread attention because it solves the limitation of single-modality imaging representation. For instance, in the realm of infrared and visible fusion \cite{CAMF,YDTR,total_variation}, infrared imaging excels in highlighting thermal targets, particularly in challenging environmental conditions, albeit at the expense of resolution. Conversely, visible imaging is adept at preserving intricate texture details but is susceptible to variations in illumination conditions. 
Consequently, the effective integration of these two imaging modalities allows their respective strengths to be fully utilized, such as enhancing visual perception and improving the identification of salient objects. Fused images exhibit improved adaptability to complex scenarios, particularly when deployed in downstream tasks such as object detection \cite{liu2022target} and semantic segmentation \cite{xie2021segformer,liu2023multi}. Moreover, image fusion finds extensive applications and extends to various computer vision tasks, including but not limited to medical image segmentation \cite{zhang2024robust,Yang,wang2024patch}, remote sensing \cite{bandara2022hypertransformer}.


In recent years, deep learning methods have become increasingly prevalent in the image fusion field. The primary challenge of image fusion lies in extracting complementary features from diverse imaging sensors and integrating them into a single image. There are three mainstream methods in multi-modal image fusion: generative methods \cite{ma2020ddcgan}, \cite{ma2019fusiongan}, \cite{xu2020mef} \cite{zhao2023mhw}, unified frameworks \cite{xu2020u2fusion}, \cite{zhang2020ifcnn}, \cite{zhang2020rethinking}, \cite{ma2022swinfusion} and auto-encoder methods \cite{li2018densefuse}, \cite{li2020nestfuse}, \cite{zhao2020didfuse}, \cite{zhao2023cddfuse}. 
It has been verified that the effective embedding strategy preserves details and enhances the overall image style \cite{yao2023laplacian}. Some multi-branch guided methods \cite{Navigating},\cite{yao2025color} have also achieved satisfactory performance when tackling specific image fusion tasks. \textbf{However, existing methods still face three challenges:} \textbf{(i)} Most methods do not fully consider the relationships between the two modal images during the image fusion training process; \textbf{(ii)} Local features extracted by convolutional neural networks (CNN) are easy to lose in the forward process, especially high-frequency information; and \textbf{(iii)} Most existing methods often overlook the facilitation of downstream tasks by the generated fused images, resulting in a lack of semantic information and salient targets. Consequently, their performance in downstream tasks is often unsatisfactory.
 
To tackle the aforementioned challenges, we propose a \textbf{S}emantic-Aware and \textbf{M}ulti-Guided Network for Infrared-Visible Image \textbf{F}usion \textbf{(SMFNet)}. \textbf{For challenge (i)}, a Graph Reasoning module is introduced to model and reason high-level relations between two modalities and extract low-level detail features. \textbf{For challenge (ii)}, a Cross Attention and Invertible Block (CAI) is employed to extract high-frequency features. And a base feature extraction module equipped with Multi-Dconv Head Transposed Attention (MDTA) and Local-enhanced Feed Forward (LeFF) network is proposed to capture global features with long-range dependencies. \textbf{For challenge (iii)}, we devise two novel loss functions for the two training stages. The first stage loss function guides the network to reconstruct source images. The second stage loss function emphasizes the preservation of target intensity and semantic textures in the fused images, intending to benefit downstream tasks. The workflow of AE-based methods, SMFNet, and quantitative results on two datasets are illustrated in Fig. \ref{1}. In this paper, our contributions are as follows:
\begin{itemize}
\item We propose a $\mathbf{S}$emantic-aware and $\mathbf{M}$ulti-guided $\mathbf{Net}$work for infrared-visible image $\mathbf{F}$usion (SMFNet) with two training stages to fully decompose features and generate fused images by fusion strategy. Graph Reason Module is used to reason high-level cross-modality relations, and also extracts respective low-level detail features.
\item The Cross Attention and Invertible Block is proposed to extract high-frequency detail features and preserve important information efficiently via learning an inverse process implicitly. The base Feature Extraction Module is applied to model long-range dependencies and obtain powerful global base features. 
\item A novel loss function based on the Gram matrix is introduced to better utilize the semantic details of intensity targets to reconstruct source images. Correlation-driven decomposition loss is proposed to trade off modality-specific, modality-shared and graph-interaction features.
\item Experiments show the SMFNet has satisfactory visual performance when compared with eleven state-of-the-art methods on $TNO$, $RoadScene$, $MSRS$, $MIF$ datasets and in two major machine perception tasks, object detection and semantic segmentation.
\end{itemize}

The remainder of this paper is organized as follows. Related works and technologies are described in Section \ref{sec2}. Section \ref{sec3_1} demonstrates the semantic-aware and multi-guided network (SMFNet). Details about the experiments and evaluation are in Section \ref{sec4}. And Section \ref{sec5} is the brief summary of the work.
\section{RELATED WORKS}
\label{sec2}
In this section, we review the existing infrared-visible image fusion algorithms in Section \ref{sec2_1}. Then, some advanced methods of Transformer and Graph Neural Networks are illustrated in Sections \ref{sec2_2} and \ref{sec2_3}.
\subsection{Infrared-Visible Image Fusion} 
\label{sec2_1}
Traditional approaches have been explored for infrared-visible image fusion in recent years. Multi-scale transform \cite{lewis2007pixel}, \cite{bhatnagar2013directive}, sparse representation-based \cite{li2020mdlatlrr}, edge-preserving filter \cite{ma2017infrared}, and saliency-based methods \cite{yang2020infrared} are widely recognized as representative methods. 
Unavoidably, traditional methods need hand-crafted fusion rules, which have restricted further development faced with numerous application scenarios.

With the progress of deep-learning technologies, researchers focus on generative methods, unified frameworks and auto-encoder methods to generate fused images for human vision and machine decision. In generative methods, Ma {\itshape et al.} \cite{ma2019fusiongan} trained the generator to generate fused images and the discriminator to force fused images similar to the source images. The denoising diffusion probabilistic model (DDPM) \cite{zhao2023ddfm} was introduced into image fusion to alleviate the unstable training for GAN-based methods. One of the representative diffusion-based works is VDMUFusion \cite{VDMU}, which integrated the fusion problem into the forward and reverse processes allowing for predicting the noise and fusion mask simultaneously. In unified frameworks, Xu {\itshape et al.} \cite{xu2020u2fusion} proposed the incorporation of elastic weight consolidation into the loss function to enable the model to handle multi-fusion tasks effectively. Zhang {\itshape et al.} \cite{zhang2020rethinking} offered an alternative approach by uniformly modeling image fusion tasks as the problem of maintaining texture and intensity proportions across source images. Evolved from traditional transforms \cite{wang2023udtcwt}, \cite{alain2014regularized}, auto-encoder methods \cite{zhao2020didfuse,zhao2023cddfuse,liang2022fusion} emphasized the encoder's role in decomposing the image into detailed features and background, with the decoder recovering the original image. Furthermore, in cooperative training methods, ReCoNet \cite{huang2022reconet} simultaneously learns networks for fusion and registration, aiming to produce robust images for misaligned source images. 
In MURF \cite{xu2023murf}, image registration and fusion mutually reinforce each other, contributing to enhanced fusion outcomes. As an end-to-end training scheme, Yao {\itshape et al.} \cite{yao2023laplacian} proposed a novel fusion network with hierarchical guidance, which obtained salient maps through Laplacian Pyramid decomposition and achieved pixel-level saliency preservation. Recently, Zheng {\itshape et al.} \cite{zheng2024probing} proposed a new paradigm based on synergistic high-order interaction in both channel and spatial dimensions. Yao {\itshape et al.} \cite{Navigating} combined the uncertainty parameters from the segmentation network to achieve enhancement and fusion process.  DLN-CFN \cite{yao2025color}, considering the content and color information, fused visible and infrared images in the low-light scenarios.

\subsection{Transformer in Image Processing} 
\label{sec2_2}
Considering the translation and rotation invariance of image properties, transformer was introduced from the field of NLP to image processing, seeking unified modeling of natural language and images. 
In 2020, Dosovitskiy {\itshape et al.} \cite{dosovitskiy2020image} proposed the Vision Transformer (ViT), which represents images as sequences of tokens, similar to how words are represented in natural language processing tasks. It has sparked significant interest and research in the application of Transformer-based architectures to computer vision (CV). Recently, subsequent works \cite{liang2021swinir, liu2022swin,xiao2022image,zhang2022swinfir,zamir2022restormer} have been proposed and applied in various CV tasks, such as image classification, high-resolution image restoration, image super-resolution, deraining and denoising. {\itshape SegFormer} \cite{xie2021segformer} comprising a hierarchical Transformer encoder without positional encoding and lightweight MLP decoders, achieved efficient segmentation on Transformers. {\itshape SwinIR} \cite{liang2021swinir}, a strong baseline for image restoration, was equipped with several residual Swin Transformer blocks together with the residual connection as the deep feature extraction module.
Wang {\itshape et al.} \cite{wang2022uformer} proposed a non-overlapping window-based self-attention and added a depth-wise convolution layer into the feed-forward network to capture local context. 
Considering the translation invariance and the vital role of local relationships, Xiao {\itshape et al.} \cite{xiao2022stochastic} proposed the stochastic window strategy to replace fixed local window strategy, which enjoys powerful representation. {\itshape XCiT} \cite{ali2021xcit} proposed cross-covariance attention, a transposed version of self-attention mechanism, what's more, the interactions between keys and queries are based on cross-covariance matrix. 
Restormer\cite{zamir2022restormer} focusing on more informative features reduced the time and memory of key-query dot-product interaction in self-attention layers. Variants of transformers also has been applied for medical and biological image processing. For medical image fusion, MM-Net \cite{liu2024mm} highlighted the strengths of the MixFormer \cite{chen2022mixformer} backbone and also fully utilized the multi-scale local features and context information.

\begin{figure*}[!htbp]
    \centering
    \includegraphics[width=\textwidth]{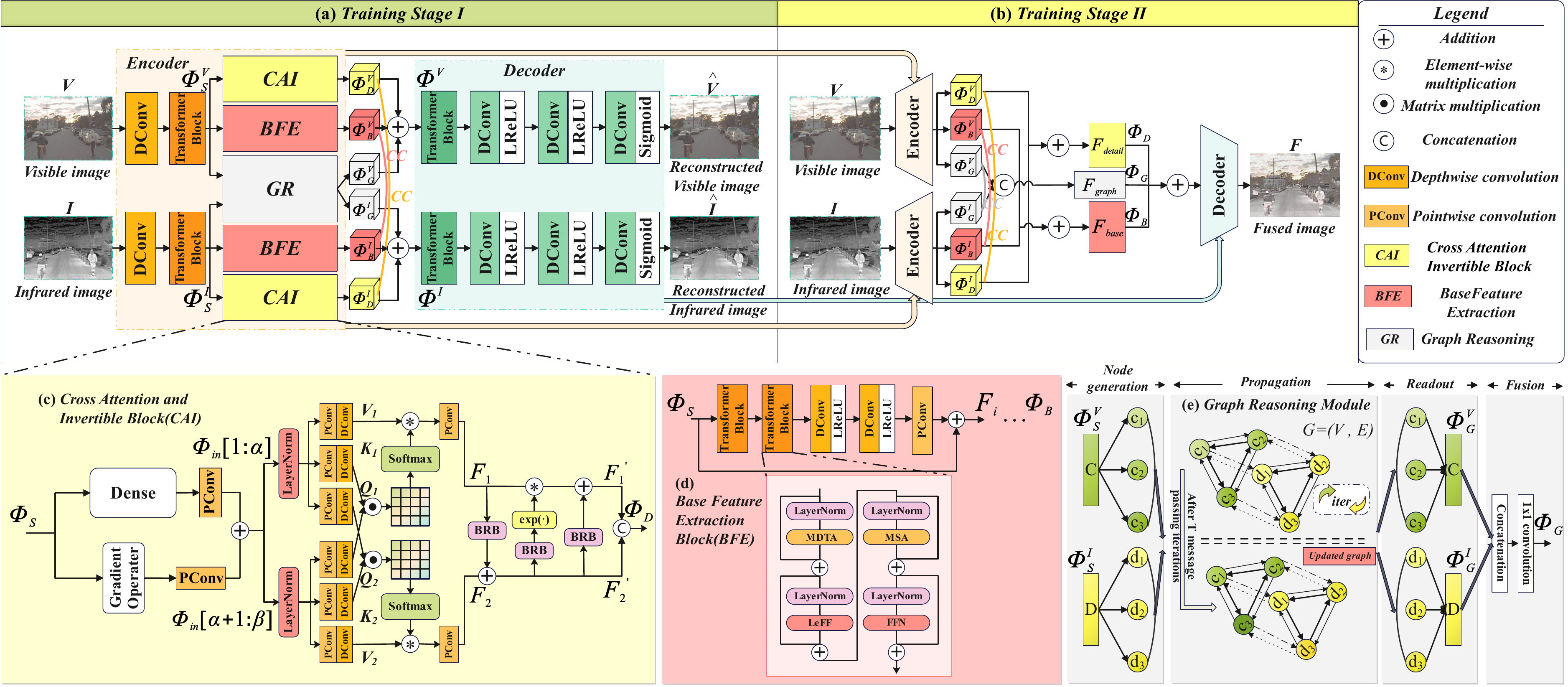}
    \caption{The architecture of proposed SMFNet. (a) In training stage I, the reconstructed original images are obtained from a three-branch encoder and a decoder framework. Features extracted from CAI, BFE and GR are aggregated and then fed into the decoder. (b) In training stage II, modality-specific, modality-shared and graph low-level features are further fused by the proposed fusion strategy. Then decoder generates the fused image. (c) CAI shows the process of refining fine-grained features and retaining high-frequency information using cross-attention mechanism and invertible module. (d) BFE enhances long-range dependencies via residual connections. (e) GR based on GNN can model the high-level relationships between two modalities shallow features and reason low-level detail features simultaneously.}
    \label{fig:overflow}
\end{figure*}

\subsection{Graph Neural Networks}
\label{sec2_3}
As a typical deep-learning network, Graph Neural Networks (GNN) \cite{MHRN} can model complex links among interpersonal relationships in social networks 
The Graph Convolutional Networks (GCN) \cite{kipf2016semi} were first proposed for semi-supervised learning on graph-structured data. As for computer vision, Wang {\itshape et al.} \cite{wang2018videos} answered the question: how do humans recognize the action "opening a book", with the definition of graph nodes by the object region from different frames in the video.
Exploring the higher performance of semantic segmentation, object detection and instance segmentation tasks, Li {\itshape et al.} \cite{li2018beyond} proposed to transform a two-dimensional image into a graph structure. The graph's vertices define clusters of pixels and the edges define the similarity between these regions in a feature space. 
Han {\itshape et al.} \cite{han2022vision} proposed Vision GNN (ViG), viewing image patches as nodes, which introduced graph convolution for aggregating and updating graph information and FFN layers for node feature transformation. 
To go a step further, {\itshape Vision HGNN} \cite{han2023vision} proposed that an image is more than a graph of nodes. They transcended conventional linkages and introduced the concept of hypergraph to encapsulate image information, which is a more universal structure.
Recently, Xu {\itshape et al.} \cite{xu2023dual} proposed a dual-space graph-based interaction network for real-time scene comprehension, which exploited cross-graph and inner-graph relations and extracted contextual information. In RGB-D salient object detection \cite{Hu24}, \cite{hu2024cross}, researchers aim at addressing a major technical challenge: how to fully leverage cross-modal complementary features. Correspondingly, GNNs have shown their abilities to learn high-level relations and describe low-level details simultaneously\cite{luo2020cascade}.

\section{Proposed Method}
\label{sec3_1}
In this section, the workflow of the proposed SMFNet is described in Section \ref{sec3_1_1}. Then, the main structure of the encoder is in Section \ref{sec3_2}. Furthermore, the proposed decoder and fusion strategy can be found in Sections \ref{sec3_3} and \ref{sec3_4}. Finally, The loss function is discussed in Section \ref{sec3_5}.
\subsection{Overview}
\label{sec3_1_1}

As illustrated in Fig. \ref{fig:overflow}, the proposed SMFNet is constructed using an encoder-decoder architecture.
Overall, \textbf{in training stage I}, the encoder's capability to extract features from the source image and the decoder's ability to obtain the reconstructed image are trained in an iterative procedure via pursuing minimal reconstruction loss; \textbf{In training stage II}, fusion layers, consistent with the encoding branches, are proposed as the fusion strategy. The identical structure of the two streams shares parameters throughout the entire training process. Constraining the similarity between the source images and the fused image in terms of structure, texture details and intensity, SMFNet can generate an image that benefits visual and computer perception; \textbf{In the inference phase}, the network structure utilized during training stage II can be directly applied to validate the efficacy of image fusion and assess the model's generalization capability in practical applications.

\subsection{Encoder}
\label{sec3_2}
\subsubsection{Shallow Feature Extraction}In this part, we mainly extract shallow features
\begin{equation}
  \mathbf{\Phi}_{S}^{V}=SFE(\mathbf{V}), \mathbf{\Phi}_{S}^{I}=SFE(\mathbf{I}).
\end{equation}
\noindent where $\mathbf{I} \in \mathbb{R}^{H\times W\times 1}$ and $\mathbf{V} \in \mathbb{R}^{H\times W\times 3}$ denote the infrared and visible images, respectively. $SFE$ represents the shallow feature extraction module, comprising a depthwise $3\times 3$ convolution and a transformer with multi-head self-attention.

\subsubsection{Cross Attention and Invertible (CAI) block} The CAI extracts high frequency information from shallow features
\begin{equation}
  \mathbf{\Phi}_{D}^{V}=CAI(\mathbf{\Phi}_{S}^{V}), \mathbf{\Phi}_{D}^{I}=CAI(\mathbf{\Phi}_{S}^{I}).
\end{equation}
 The main structure of CAI is shown in Fig. \ref{fig:overflow} (c). \textbf{Firstly}, to extract fine-grained detail information, we introduce a gradient and residual dense block benefiting from the Sobel operator. The block can be represented as
\begin{equation}
    \mathbf{\Phi}_{in} = PConv(Dense((\Phi_S))+PConv(\nabla(\Phi_S)),
\end{equation}
\noindent where $\Phi_S$ indicates infrared/visible shallow features extracted by SFE, $\nabla$ denotes the sobel gradient operator. As the variant of resblock, dense connection makes full use of features extracted from convolutions and activate functions. Gradient magnitude is calculated to remove channel dimension differences.

\begin{figure}[h]
    \centering
    \includegraphics[width=\linewidth]{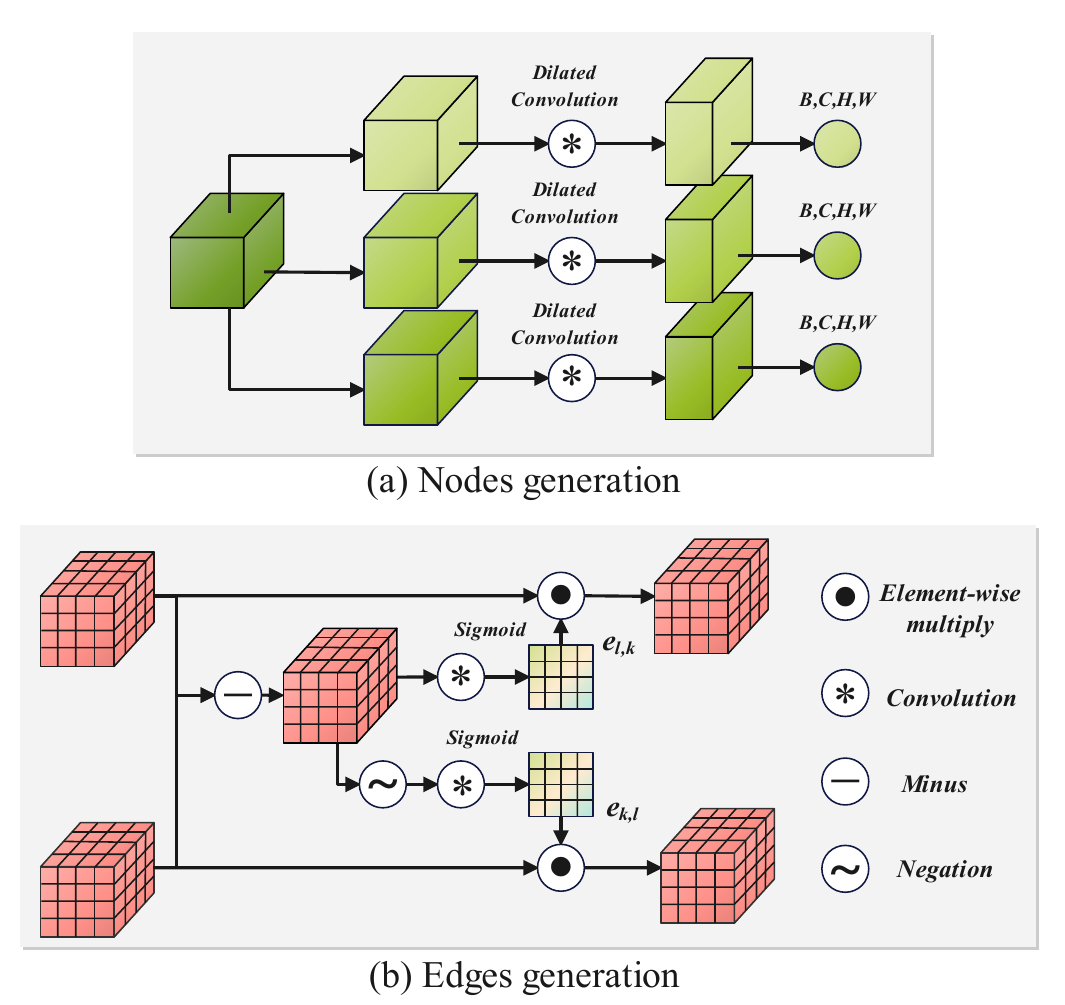}
    \caption{Specific illustration for nodes and edges generating process in graph reasoning module.}
    \label{generation}
\end{figure}

\textbf{Secondly}, to aggregate the information from CNN's dual branches through interactive cross-attention mechanism and information invertibility, we propose a cross-attention module with the ability of local feature extraction. We set $\mathbf{\Phi}_{in}[1:\alpha]$ and $\mathbf{\Phi}_{in}[\alpha+1:\beta]$ separated from $\mathbf{\Phi}_{in}[1:\beta]$ as intermediate features used for the dual-branch block. $ \mathbf{Q}_1$, $\mathbf{Q}_2 \in \mathbb{R}^{H\times W\times C}$, $\mathbf{K}_1$, $\mathbf{K}_2 \in \mathbb{R}^{C\times H\times W}$, 
$\mathbf{V}_1$, $\mathbf{V}_2 \in \mathbb{R}^{H\times W\times C}$. \{$ \mathbf{Q}_1$, $\mathbf{K}_1$, $\mathbf{V}_1$\} and \{$ \mathbf{Q}_2$, $\mathbf{K}_2$, $\mathbf{V}_2$\} are obtained after $\mathbf{\Phi}_{in}[1:\alpha]$ and $\mathbf{\Phi}_{in}[\alpha+1:\beta]$ pass through the layer norm, pointwise convolution and depthwise convolution, respectively. The cross-attention module can be represented as

\begin{equation}
\mathbf{F}_{1}=\mathbf{V}_1*Softmax(\mathbf{K}_1,\mathbf{Q}_2/T),
  \label{eq1}
\end{equation}
\begin{equation}
\mathbf{F}_{2}=\mathbf{V}_2*Softmax(\mathbf{K}_2,\mathbf{Q}_1/T).
\label{eq2}
\end{equation}

\noindent where $T$ means temperature, a learnable scaling parameter. The cross-attention module can divide the number of channels to the head, which achieves two branches of learning attention maps simultaneously.

\textbf{Finally}, to refine the feature extracted from the above cross-attention module, and efficiently obtain semantic feature as expected, we learn an inverse process implicitly  \cite{zhou2022mutual,jing2021hinet,guan2022deepmih,zhou2024general}. The intermediate features are formulated as
\begin{equation}
    \mathbf{F}_{1}^{'}=\mathbf{F}_{1} \otimes exp(\psi(\mathbf{F}_{2})) +\psi(\mathbf{F}_{2}),   
\end{equation}
\begin{equation}
    \mathbf{F}_{2}^{'}=\mathbf{F}_{2}+\psi(\mathbf{F}_{1}),
\end{equation}
\noindent where $\otimes$ denotes the element-wise multiplication, $\psi(\cdot)$ refers to the Bottleneck Residual Block (BRB), and the $exp(\cdot)$ indicates the exponential function. The output of CAI $\mathbf{\Phi}_{D}$ can be obtained by the concatenation of $\mathbf{F}_{1}^{'}$ and $\mathbf{F}_{2}^{'}$.

\subsubsection{Base Feature Extraction (BFE) Module} Contrary to CAI, the BFE extracts low-frequency global features from SFE-extracted shallow features
\begin{equation}
  \mathbf{\Phi}_{B}^{V}=BFE(\mathbf{\Phi}_{S}^{V}), \mathbf{\Phi}_{B}^{I}=BFE(\mathbf{\Phi}_{S}^{I}).
\end{equation}

The main structure of BFE is in Fig. \ref{fig:overflow} (d). Using Multi-Dconv Head Transposed Attention (MDTA), the transformer implements channel-dimension self-attention rather than the spatial dimension. It computes the cross-covariance among channels to generate an attention map to model global features with linear complexity. Local-enhanced Feed Forward (LeFF) Network, adding depth-wise convolution to Feed Forward Network
(FFN) aims to reduce limited capability of leveraging local context. BFE can be defined as follows in detail

\begin{figure}[h]
    \centering
    \includegraphics[width=\linewidth]{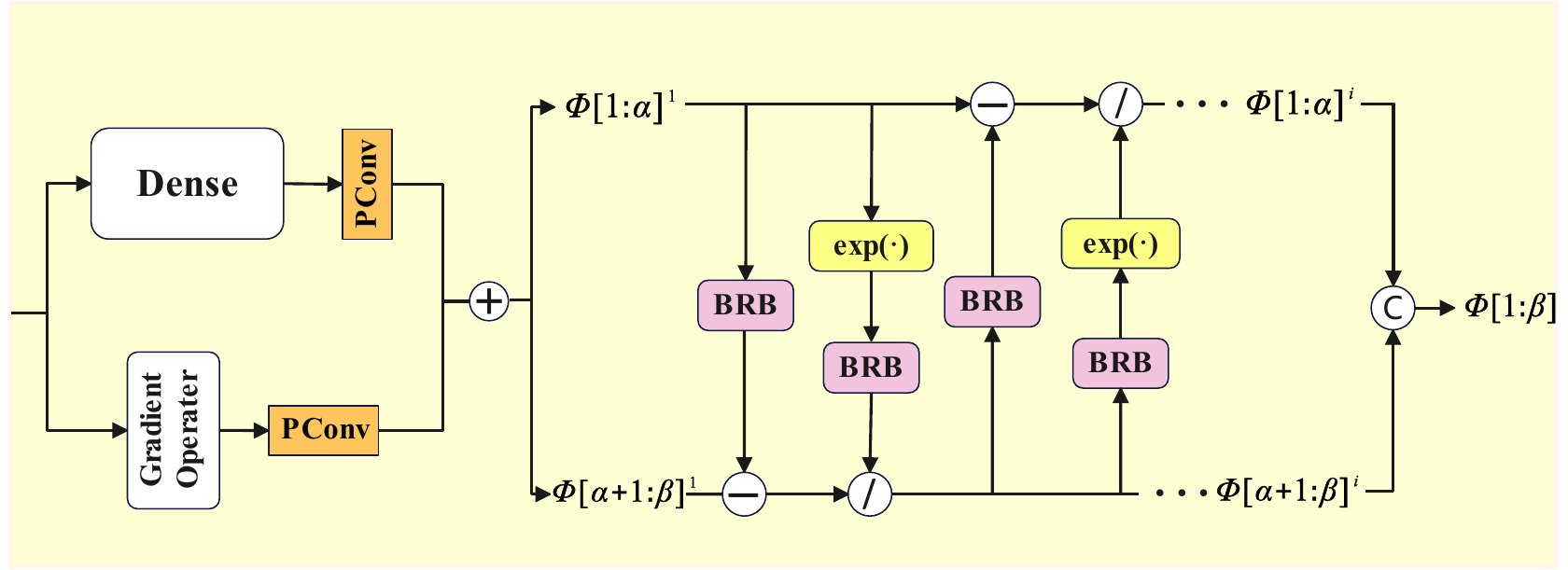}
    \setlength{\abovecaptionskip}{-0.1cm}   
    \caption{Proposed detail fusion layers in training stage II.}
    \label{detail_fusion_layer}
\end{figure}

\begin{equation}
    \mathbf{F}_{0}=\mathbf{\Phi}_{S},
\end{equation}
\begin{equation} \mathbf{F}_{1}=Conv(\rho(\rho(\mathbf{F}_{0})))+\mathbf{F}_{0}, 
\end{equation}
\begin{equation}
\mathbf{\Phi}_{B}=Conv(\rho(\rho(\mathbf{F}_{1})))+\mathbf{F}_{1},
\end{equation}

\noindent where $\rho(\cdot)$ is the residual BFE block. The residual connection promotes feature aggregation from different feature levels during the phase of establishing long-range dependencies.


\subsubsection{Graph Reasoning (GR) Module}
The GR module is the third branch in the encoder, and it can be represented as
\begin{equation}
  \mathbf{\Phi}_{G}^{V}=GR(\mathbf{\Phi}_{S}^{V}), \mathbf{\Phi}_{G}^{I}=GR(\mathbf{\Phi}_{S}^{I}).
\end{equation}

As shown in Fig. \ref{fig:overflow} (e), the GR module is composed of node embedding, 
edge embedding, information delivery and node update, 
aiming at hierarchically reason about relations between two modalities and exploring useful information.
\label{sec3}

{\itshape Graph Creation.} We set $\mathbf{C}=[c_1,c_2,...,c_n]$ and 
$\mathbf{D}=[d_1,$
$d_2,...,d_n]$ 
as nodes from visible and infrared, two kinds of modalities with $n$ scales. $\mathbf{G}=(\mathbf{V},\mathbf{E})$ stands for a directed graph. The inner nodes $V={\mathbf{C}}\cup{\mathbf{D}}$ . When starting a graph, $c_i\leftrightarrow c_j$  and $d_i\leftrightarrow d_j$ belong to the same modality but different scales, $c_i\leftrightarrow d_i$ are the same scale but different modalities meanwhile. $N=2 \cdot n$ and $N(l)$ stands for the number of nodes totally in graph $G$. The generation process of nodes and edges is illustrated in Fig. \ref{generation} (a) and (b).

{\itshape Multi-scale Nodes Embedding.} Each scale node has the shape $c_i^{(0)}$, $d_i^{(0)} \in \mathbb{R}^{H\times W \times C}$. For example, for the current nodes $c_{i} \in \mathbf{C}$ generation
\begin{equation}
    \mathbf{c}_i^{(0)}=Conv(Pool({c}_i^{(0)})),
\end{equation}

\noindent where $Pool(\cdot)$ implements triple branches pooling and convolution operations.

{\itshape Edges Embedding.} Assume that $v_k,v_l \in V$, including nodes from visible/infrared modalities $c_i$ and $d_i$, therefore, we obtain directed edges generation scheme from $v_k$ to $v_l$
\begin{equation}
    e_{k,l}=Conv(v_l-v_k).
\end{equation}

{\itshape Information delivery.} After nodes and edges generation, 
information is transferred from the last node to the current one via powerful edges. We set $m_{k,l}^{(t)}$ as message from $v_k$ to $v_l$ via $e_{k,l}$ in $t-1$ state while delivering

\begin{equation}
    m_{k,l}^{(t)}=\underset{k \in N(l)}{\Sigma}Sigmoid(e_{k,l}^{(t-1)}) \cdot v_k^{(t-1)}.
\end{equation}

{\itshape Nodes Update.} Next to the message delivery, GR module learns intricate cross-modality relations, and nodes are updated through a GRU mechanism so that we obtain the updated node as $v^{(t)}$
\begin{equation}
    v_l^{(t)}=\underset{k \in N(l)}{\Sigma}F_{GRU}(v_l^{(t-1)},m_{k,l}^{(t-1)}).
\end{equation}
Then, $C$ and $D$ can be merged separately through the convolution operation.

\subsection{Fusion Strategy}
\label{sec3_3}
The functions of $F_{detail}$, $F_{base}$, $F_{graph}$ are fusion layers to fuse detail, base and graph reasoning features, respectively.
\begin{equation}
  \mathbf{\Phi}_{D}=F_{detail}(\mathbf{\Phi}_{D}^{V}, \mathbf{\Phi}_{D}^{I}),
  \mathbf{\Phi}_{B}=F_{base}(\mathbf{\Phi}_{B}^{V}, \mathbf{\Phi}_{B}^{I}),
  \end{equation}

\begin{equation}
  \mathbf{\Phi}_{G}=F_{graph}(\mathbf{\Phi}_{G}^{V}, \mathbf{\Phi}_{G}^{I}).
\end{equation}
Detail fusion layers are shown in Fig. \ref{detail_fusion_layer}. We set $Output(\cdot)$ as the output of the one branch. Mathematical expressions are as follows:
\begin{equation}
\label{eq7}
\begin{aligned}
    Output(\mathbf{\Phi}[1:\alpha]^{i+1})=(\mathbf{\Phi}[1:\alpha]^{i}- \psi(\mathbf{\Phi} \\ [\alpha+1:\beta]^{i})) \otimes exp(-\alpha(\psi(\mathbf{\Phi}[\alpha+1:\beta]^{i}))),    
\end{aligned}
\end{equation}
\begin{equation}
\label{eq8}
\begin{aligned}
    Output(\mathbf{\Phi}[\alpha+1:\beta]^{i+1})=(\mathbf{\Phi}[\alpha+1:\beta]^{i}-\\ \psi(\mathbf{\Phi}[1:\alpha]^{i})) \otimes exp(-\alpha(\psi(\mathbf{\Phi}[1:\alpha]^{i}))).
\end{aligned}
\end{equation}
The $i$th layer designed for fusing cross-modality detail feature is shown as Eqs. (\ref{eq7}) and (\ref{eq8}). Eventually, the last two layers' features are concatenated into the decoder part. As described in Fig. \ref{fig:overflow} (e), $\mathbf{\Phi}_V^G$, $\mathbf{\Phi}_I^G$ from visible and infrared modalities are concatenated into a pointwise $1 \times 1$ convolution to fuse the output of read out part.

In consistency with the base feature extraction structure in the encoder, the base fusion layer is composed of a transformer the same as the block in $SFE$ with MDTA and LeFF aiming at fusing global features $\mathbf{\Phi}_B^V$ and $\mathbf{\Phi}_B^I$.

\subsection{Decoder}
\label{sec3_4}
The decomposed features are aggregated together into a decoder to output original images (in stage I) or fused images (in stage II).

\textbf{In stage I}, the outputs of the decoder, reconstructed original images are 
\begin{equation}
  \mathbf{\hat{I}}=D(\mathbf{\Phi}_{D}^{I}, \mathbf{\Phi}_{B}^{I}, \mathbf{\Phi}_{G}^{I}),
  \mathbf{\hat{V}}=D(\mathbf{\Phi}_{D}^{V}, \mathbf{\Phi}_{B}^{V}, \mathbf{\Phi}_{G}^{V}).
  \end{equation}

\textbf{In stage II}, the output of the decoder, fused image is 
\begin{equation}
  \mathbf{F}=D(\mathbf{\Phi}_{D}, \mathbf{\Phi}_{B}, \mathbf{\Phi}_{G}).
  \end{equation}
The transformer block within the decoder adopts an identical structure to those in the $SFE$ and the base fusion layer. It utilizes depthwise convolution to reconstruct aggregated features from the space $\mathbb{R}^{B \times C \times H \times W}$ to $\mathbb{R}^{H \times W \times 1}$. Then the fused image  $\mathbb{R}^{H \times W \times 3}$ is obtained via YCbCr space conversion \cite{zhao2023cddfuse}. The details of the whole network are listed in Tab. \ref{network}.

\begin{table}
\centering
\small
\caption{Details architecture of our proposed SMFNet. Head refers to multi-head self-attention in a Transformer model. The terms I, O, and K correspond to the input dimension, output dimension, and kernel size, respectively.}
\label{network}
\resizebox{0.8\linewidth}{!}{
\begin{tabular}{c | c | c}
    \toprule
      Layer & Explanation & Module \\ \midrule
      
        \multirow{5}{*}{\centering Encoder} & \multirow{2}{*}{\centering Fig. 2(a)} & DConv(I1O64K3) \\ 
        ~ & ~ & Transformer(I64O64Head8) \\ 
        ~ & Fig. 2(c) & CAI Block(I64O64) \\ 
        ~ & Fig. 2(d) & BFE Block(I64O64) \\ 
        ~ & Fig. 2(e) & GR Module(I64O64) \\ \midrule
        $F_{detail}$ & Fig. 4 & Detail Module(I64O64) \\ 
        $F_{base}$ & - & Transformer(I64O64Head8) \\ 
        $F_{graph}$ & - & Conv(I128O64K1) \\ \midrule
        \multirow{4}{*}{\centering Decoder} & \multirow{4}{*}{\centering Fig. 2(a)} & Transformer(I64O64Head8) \\ 
        ~ & ~ & DConv(I64O64K3) \\ 
        ~ & ~ & DConv(I64O64K3) \\ 
        ~ & ~ & DConv(I64O64K3), Sigmoid \\ 
    \bottomrule
\end{tabular}
}
\centering
\end{table}

\subsection{Loss Function}
\label{sec3_5}
\textbf{In stage I}, we train a reconstruction network with two-fold objectives: it should output reconstructed images without any loss of visual detail and also decompose the input images into base and detail layers. Therefore, the total loss is formulated as 

\begin{equation}
\label{25}
  \mathcal{L}_{total}^I= \underbrace{\mathcal{L}_{vi}+\mathcal{L}_{ir}}_{Reconstruction \enspace loss}+\alpha_1 \underbrace{\mathcal{L}_{decomp}^{I}}_{Decomposition\enspace loss}, 
\end{equation}
 
\noindent where $\mathcal{L}_{vi}$ and $\mathcal{L}_{ir}$ are the reconstruction losses for visible and infrared images, respectively. $\mathcal{L}_{decomp}^I$ denotes the feature decomposition loss. $\alpha_1$ is a balancing factor. Both the two reconstruction losses $\mathcal{L}_{vi}$ and $\mathcal{L}_{ir}$ have structural loss $\mathcal{L}_{SSIM}$ and gradient loss $\mathcal{L}_{grad}$. $\mathcal{L}_{vi}$ is defined as
\begin{equation}
\label{26}
  \mathcal{L}_{vi}=\beta_1\mathcal{L}_{SSIM}(\mathbf{V}, \mathbf{\hat{V}})+\beta_2\mathcal{L}_{grad}(\mathbf{V}, \mathbf{\hat{V}}),
\end{equation}
\noindent where the gradient
loss is defined as
$\mathcal{L}_{grad}
(\mathbf{V},\mathbf{\hat{V}})=||\left|\nabla V\right|-\left|\nabla\hat{V}\right|||_1$ ($\nabla$ denotes the gradient operator), and the structural loss is defined as $\mathcal{L}_{SSIM}(\mathbf{V}, \mathbf{\hat{V}})=
1-SSIM(\mathbf{V}, \mathbf{\hat{V}})$. $SSIM$ is the structural similarity index, which measures the similarity of two images in terms of luminance, contrast, and structure \cite{wang2004image}. 

In Eq. (\ref{25}), the reconstruction loss for the infrared image is defined as 
\begin{equation}
\label{27}
  \mathcal{L}_{ir}=\mathcal{L}_{semantic}(\mathbf{I},\mathbf{\hat{I}})+\beta_1\mathcal{L}_{SSIM}(\mathbf{I},\mathbf{\hat{I}})+\beta_2\mathcal{L}_{grad}(\mathbf{I},\mathbf{\hat{I}}),
\end{equation}

\noindent where $\mathcal{L}_{grad}(I,\hat{I})$
and $\mathcal{L}_{SSIM}(I,\hat{I})$
in $\mathcal{L}_{ir}$ can be obtained in the same way as the ones in $\mathcal{L}_{vi}$, and $\beta_1, \beta_2$ are set the same as Eq. (\ref{26}). As infrared images contain minimal detailed information about the background but offer a clear display of object shapes and contours, the reconstructed infrared images should retain the crucial and salient information of the original images. Therefore, we define the $\mathcal{L}_{semantic}$ as

\begin{equation}
  \mathcal{L}_{semantic}(\mathbf{I},\mathbf{\hat{I}})=||\mathbf{Gr(I)},\mathbf{Gr(\hat{I})}||_2,
\end{equation}
\noindent where $\mathbf{Gr}(\cdot)$ represents Gram matrix. The reason why the matrix is employed is that it contains the semantic features of images  \cite{li2023lrrnet}.

According to decomposed features from the encoder, we assume that modality-specific information has a lower value of correlation coefficient and modality-shared information has a higher correlation value. $\mathcal{L}_{decomp}^I$ is formulated as
\begin{equation}
    \mathcal{L}_{decomp}^I=\dfrac{(CC(\Phi_{D}^{\mathbf{V}}, \Phi_{D}^{I}))^2}{CC(\Phi_{B}^{V}, \Phi_{B}^{I})+\delta},
\end{equation}
\noindent where $CC(\cdot)$ denotes the correlation coefficient. $\delta$ is set to 1.01, which can guarantee gradient descent and optimization.

\textbf{In stage II}, the total loss is
\begin{equation}
    \label{34}
\mathcal{L}_{total}^{II}=\mathcal{L}_{intensity}^{II}+\alpha_2\mathcal{L}_{grad}^{II}+\alpha_3\mathcal{L}_{decomp}^{II},
\end{equation}
\noindent where $\mathcal{L}_{intensity}^{II}$ and $\mathcal{L}_{grad}^{II}$ are inspired by \cite{tang2022image},  building the restriction among the fused image $\mathbf{F}$ and the two source images $\mathbf{V}$ and $\mathbf{I}$ by
\begin{equation}
  \mathcal{L}_{intensity}^{II}=||\mathbf{F}-\max{(\left|\mathbf{I}\right|,\left|\mathbf{V}\right|)}||_1,
\end{equation}
\begin{equation}
  \mathcal{L}_{grad}^{II}=||\left|\nabla \mathbf{F}\right|-\max{(\left|\nabla \mathbf{I}\right|,\left|\nabla \mathbf{V}\right|)}||_1.
\end{equation}

Different from $\mathcal{L}_{decomp}^{I}$, high-frequency information shows more texture and details, whereas low-frequency information emphasizes background and base features. $CC(\Phi_{G}^{V}, \Phi_{G}^{I})$ is added to balance features extracted from other branches and also as the complementary for simple modality-specific features
\begin{equation}
\label{eq27}
    \mathcal{L}_{decomp}^{II}=\dfrac{(CC(\Phi_{D}^{V}, \Phi_{D}^{I}))^2+(CC(\Phi_{G}^{V}, \Phi_{G}^{I}))^2}{CC(\Phi_{B}^{V}, \Phi_{B}^{I})+\delta}.
\end{equation}


\begin{figure*}[h]
    \centering
    \includegraphics[width=\textwidth]{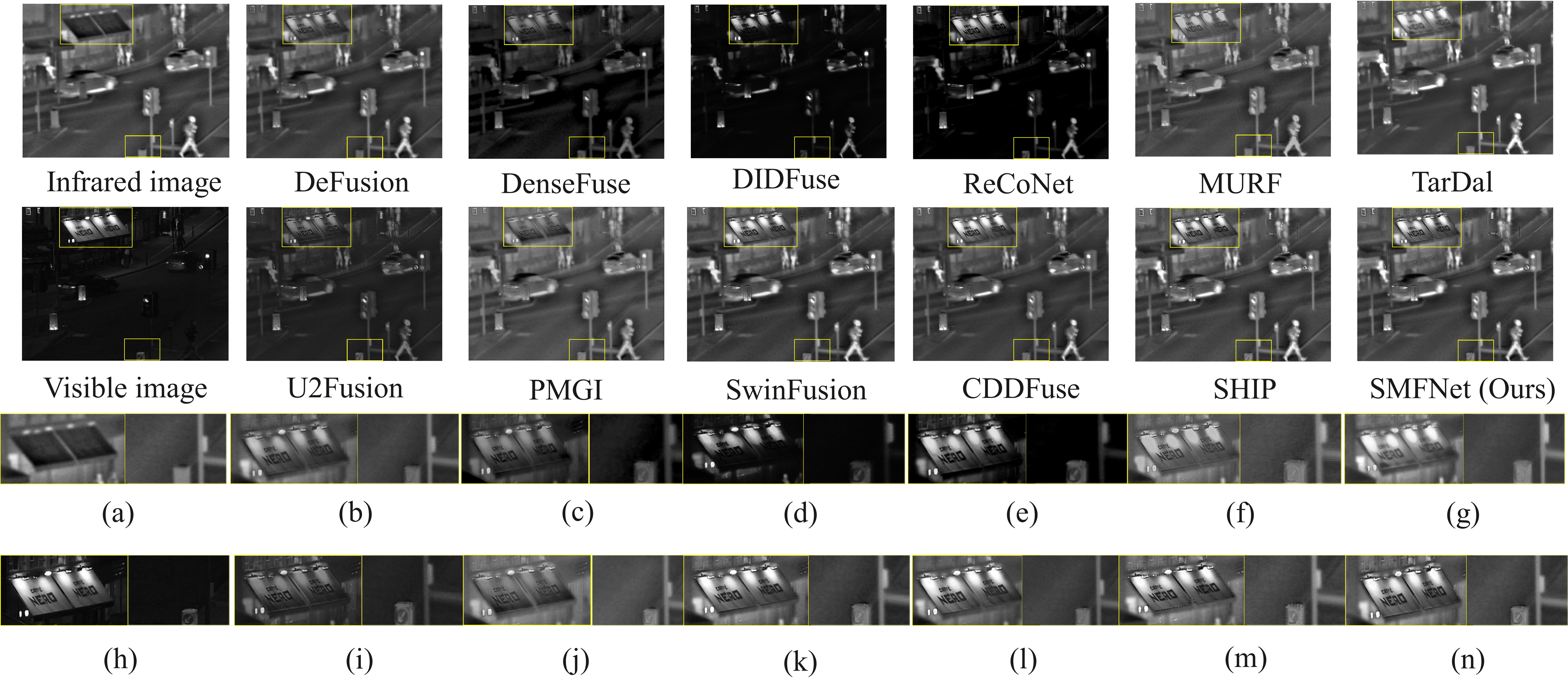}
    \caption{Qualitative fusion results on TNO \cite{toet2012progress} dataset. Magnified areas from (a) to (n) show the performance of the shade board on the eave and the traffic sign arrow pointing downwards to the left, corresponding to the visible, infrared, and fused images generated by the state-of-the-art methods.}
    \label{TNO}
\end{figure*}
\begin{figure*}[h]
    \centering
    \includegraphics[width=\textwidth]{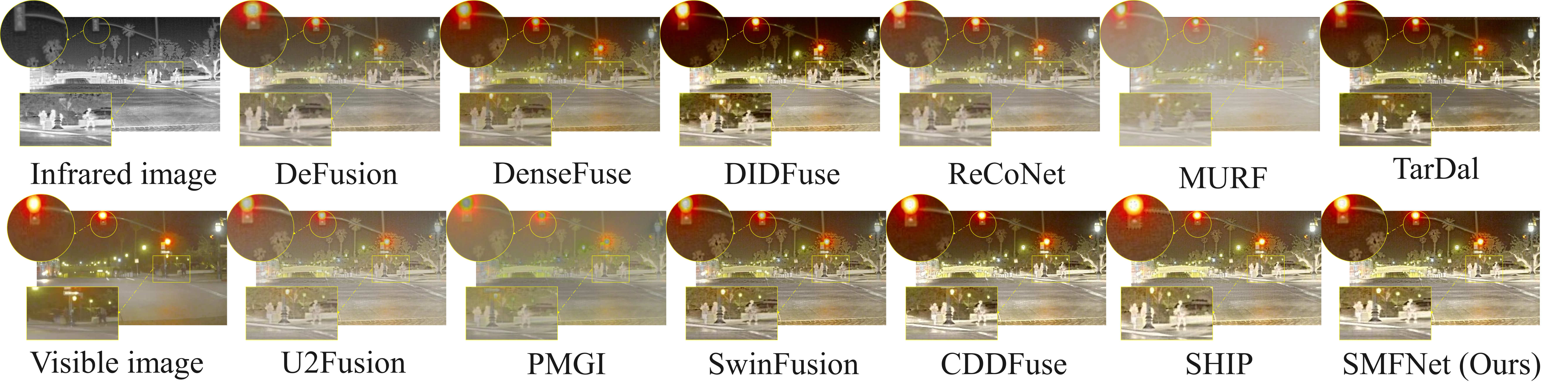}
    \caption{Qualitative fusion results on RoadScene \cite{xu2020u2fusion} dataset.}
    \label{RoadScene}
\end{figure*}
\begin{figure*}[h]
    \centering
    \includegraphics[width=\textwidth]{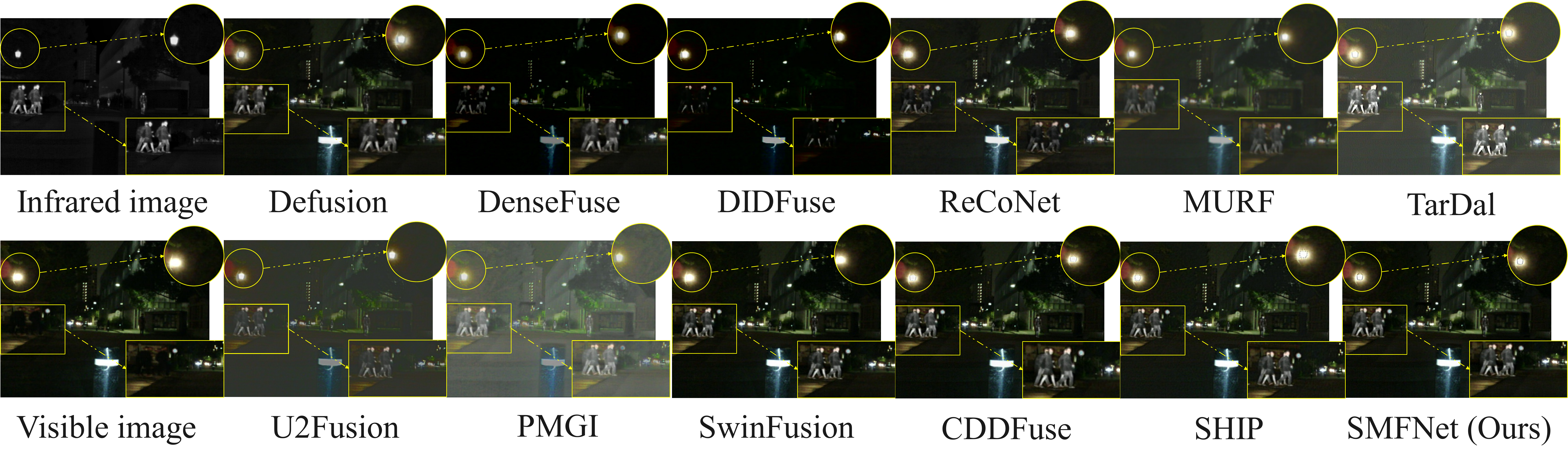}
    \caption{Qualitative fusion results on MSRS \cite{tang2022piafusion} dataset.}
    \label{MSRS}
\end{figure*}




\section{Experiments}
\label{sec4}
In this section, datasets and experimental settings are first described in Sections \ref{sec4_1} and \ref{sec4_2}. Subsequently, we conduct experiments to justify the superiority of the proposed algorithm by answering the following six questions: 
\begin{itemize}
\item $\mathbf{RQ1}$: How does the proposed method perform on the mainstream infrared-visible image fusion datasets?
\item $\mathbf{RQ2}$: Whether the proposed method generates task-oriented fused images to facilitate downstream tasks compared to other SOTA fusion algorithms?
\item $\mathbf{RQ3}$: How does the proposed model perform on other types of image fusion?
\item $\mathbf{RQ4}$: How do the proposed structures in the encoding network and fusion strategy influence the performance of the proposed model?
\item $\mathbf{RQ5}$: What role does the loss function proposed in this article play in guiding the neural network training process?
\item $\mathbf{RQ6}$: Why do we have two training stages?
\end{itemize}

\subsection{Experiment Configuration}
\label{sec4_1}
In the experimental setup, we partitioned the dataset into three distinct subsets for training, validation, and testing purposes. Specifically, we employed 1083 image pairs from \cite{tang2022piafusion} for training, while 50 RoadScene pairs served as the validation set. For testing, we utilized a combination of datasets, including 25 pairs from TNO \cite{toet2012progress}, 50 pairs from RoadScene \cite{xu2020u2fusion}, and 361 pairs from MSRS (Multi-Spectral Road Scenarios) \cite{tang2022piafusion}. Although our model supports inputs of arbitrary size, we ensure that each iteration receives the same amount of data. During the training phase, the MSRS images were initially converted into grayscale and then cropped into patches of dimensions $128 \times 128$. The training process encompassed a total of 120 epochs, with the initial 40 epochs designated as the first training stage, followed by an additional 80 epochs constituting the second stage. The batch size was fixed at 6 throughout the training procedure. 

As for hyperparameters in Eqs. (\ref{25}), (\ref{26}), (\ref{27}) and (\ref{34}), $\alpha_{1}$ was uniformly set to 2; $\beta_{1}$ and $\beta_{2}$ were uniformly set to 8 and 10 in training stage I; $\alpha_{2}$ and $\alpha_{3}$ were set to 10 and 2 in training stage II. We used the Adam optimizer, and the initial learning rate was ${1e}^{-4}$ without weight decay. Experiments were executed on a PC with two NVIDIA GeForce 4090 GPUs and the pytorch framework.

\subsection{Comparision Settings}
\label{sec4_2}
Eight evaluation metrics were employed to objectively measure the quality of fused images, including Entropy (EN), standard deviation (SD), spatial frequency (SF), mutual information (MI), sum of correlations of differences (SCD), visual information fidelity (VIF), Qabf, average gradient (AG), and structural similarity (SSIM). The larger values of the above metrics mean better image quality. Detailed descriptions of these metrics can be referenced in \cite{zhao2023cddfuse, ma2019infrared}. We compared the proposed model with eleven state-of-the-art methods: DeFusion \cite{liang2022fusion}, DenseFuse \cite{li2018densefuse}, DIDFuse \cite{zhao2020didfuse}, MURF \cite{xu2023murf}, ReCoNet \cite{huang2022reconet}, SwinFusion \cite{ma2022swinfusion}, TarDAL \cite{liu2022target}, U2Fusion \cite{xu2020u2fusion}, PMGI \cite{zhang2020rethinking}, CDDFuse \cite{zhao2023cddfuse} and SHIP \cite{zheng2024probing}. We directly run the vanilla codes released by the authors on the three test datasets.

\subsection{Infrared-Visible Fusion Results ($\mathbf{RQ1}$)}
\subsubsection{Qualitative comparison} 
In this section, we show qualitative results on TNO \cite{toet2012progress}, Roadscene \cite{xu2020u2fusion} and MSRS \cite{tang2022piafusion} datasets in Fig. \ref{TNO}, \ref{RoadScene} and \ref{MSRS}, respectively. Through thorough observation and comparison, our method embodies satisfactory visual performance compared with other SOTA methods. For example, targets in dark environments can be highlighted, such as people and cars. Furthermore, a clear background in the visible scene allows for the display of rich texture details. Notably, with attention to lighting, we can discern clear shape and contour information.

\subsubsection{Quantitative comparison} In Tab. \ref{tab:VIF}, we present eight evaluation metrics across three test sets for quantitative analysis. Specifically, the highest EN and AG indicate significant preservation of information from the source images in our fused images, thanks to the cross-attention invertible module and effective complementary feature aggregation. The attention to targeted edge contours is largely attributed to the semantic loss function. Additionally, the highest values of VIF and Qabf indicate that the generated fused images exhibit excellent visual performance suitable for the human visual system and are also well-suited for subsequent computer vision tasks. Furthermore, the highest SF demonstrates good texture details in our results.

\begin{table*}
\centering
\caption{Quantitative evaluation results of IVF task, where \textcolor{red}{\textbf{red}} indicates the best results, and \textcolor{blue}{\textbf{blue}} shows the second-best values.}
\label{tab:VIF}
\resizebox{\textwidth}{!}{
\begin{tabular}{ l | c  c  c  c  c  c  c  c | c  c  c  c  c  c  c  c | c  c  c  c  c  c  c  c  c  c  }
    \toprule
     \multicolumn{1}{c|}{\multirow{2}{*}{\textbf{Method}}} & \multicolumn{8}{c|}{\textbf{TNO Infrared-Visible Fusion Dataset}} &
    \multicolumn{8}{c|}{\textbf{RoadScene Infrared-Visible Fusion Dataset}} &
     \multicolumn{8}{c}{\textbf{MSRS Infrared-Visible Fusion Dataset}}\\
& EN &  SD &  SF &  MI & VIF & Qabf &  AG & SSIM & EN &  SD &  SF &  MI & VIF & Qabf &  AG & SSIM & EN &  SD &  SF &  MI & VIF & Qabf &  AG & SSIM\\ \midrule
        DeFusion\cite{liang2022fusion} & 6.5822 & 30.9861 & 6.5973 & 1.7573 & 0.5528 & 0.3589 & 3.7822 & \textcolor{red}{\textbf{1.4246}}
        & 6.8545 & 34.4052 & 7.7822 & 2.1668 & 0.5746 & 0.4129 & 4.3117 & \textcolor{blue}{\textbf{1.4486}}
        & 6.3825 & 35.4282 & 8.1468 & 2.0686 & 0.7304 & 0.5067 & 3.7396 & \textcolor{red}{\textbf{1.4400}}
        \\ 
        DenseFuse\cite{li2018densefuse} & 6.2963 & 27.1425 & 7.0023 & 1.4947 & 0.5519 & 0.3394 & 3.7325 & 1.3891 
        & 6.6234 & 28.5042 & 7.4016 & 1.985 & 0.562 & 0.3688 & 3.9822 & \textcolor{red}{\textbf{1.4716}}
        & 4.7287 & 26.2808 & 6.3539 & 1.4026 & 0.4298 & 0.2899 & 2.6237 & 0.8176
        \\ 
        DIDFuse\cite{zhao2020didfuse} & 6.8621 & \textcolor{red}{\textbf{46.8716}} & 11.7683 & 1.6953 & 0.593 & 0.4033 & 6.0095 & 1.1797 
        & 7.3207 & \textcolor{blue}{\textbf{52.8861}} & 13.6309 & 2.0969 & 0.6215 & 0.4855 & 7.3586 & 1.2839
        & 4.5002 & 29.6771 & 9.6158 & 1.4373 & 0.3054 & 0.2038 & 2.9108 & 0.8365
        \\ 
        ReCoNet\cite{huang2022reconet} & 6.6775 & 40.4573 & 7.958 & 1.7181 & 0.5307 & 0.3728 & 4.7424 & 1.2751
        & 7.1903 & 46.5624 & 9.753 & 2.1219 & 0.5491 & 0.3848 & 5.5852 & 1.3351
        & 6.501 & 40.0413 & 9.0682 & 2.1484 & 0.6927 & 0.4913 & 4.5338 & 1.3155
        \\ 
        MURF\cite{xu2023murf} & 6.5402 & 30.7797 & 12.2408 & 1.6282 & 0.5566 & 0.4286 & 6.3707 & 1.3367 
        & 6.1869 & 21.6954 & 7.1072 & 1.4761 & 0.2033 & 0.1562 & 3.6145 & 1.2126
        & 6.1556 & 28.5848 & 4.7218 & 1.6651 & 0.4269 & 0.1583 & 2.4601 & 1.0879
        \\ 
        TarDal\cite{liu2022target} & 6.7795 & 40.6012 & 11.4994 & 1.9477 & 0.5677 & 0.4113 & 5.8702 & 1.3149 
        & 7.0848 & 42.7459 & 11.1432 & \textcolor{blue}{\textbf{2.4321}} & 0.5826 & 0.4386 & 5.7274 & 1.3677
        & 6.3476 & 35.4603 & 9.8727 & 1.8341 & 0.6728 & 0.4255 & 4.3889 & 1.0186
        \\ 
        U2Fusion\cite{xu2020u2fusion} & 6.4175 & 26.337 & 8.8514 & 1.3468 & 0.5391 & 0.4251 & 5.0897 & \textcolor{blue}{\textbf{1.4120}}
        & 6.7108 & 29.507 & 10.2725 & 1.8396 & 0.5599 & 0.4914 & 5.7803 & 1.4350
        & 4.9532 & 18.869 & 6.7124 & 1.3533 & 0.4742 & 0.315 & 2.9513 & 1.0379
        \\ 
        PMGI\cite{zhang2020rethinking} & 6.9519 & 35.832 & 8.9051 & 1.6512 & 0.6034 & 0.4119 & 5.0468 & 1.3509 
        & 6.7964 & 31.9394 & 7.5574 & 2.0768 & 0.5442 & 0.3591 & 4.333 & 1.3611
        & 5.8333 & 18.1081 & 6.0163 & 1.3504 & 0.6175 & 0.2731 & 3.0761 & 0.8686
        \\ 
        SwinFusion\cite{ma2022swinfusion} & 6.8998 & 41.1725 & 11.372 & \textcolor{blue}{\textbf{2.3132}} & 0.7572 & 0.5381 & 6.1813 & 1.3771
        & 6.8274 & 31.7477 & 12.8767 & 1.7429 & 0.6093 & \textcolor{red}{\textbf{0.584}} & 6.8742 & 1.4344
        & 6.6185 & \textcolor{blue}{\textbf{42.9849}} & 11.0554 & \textcolor{blue}{\textbf{3.3303}} & 1.0056 & 0.6693 & 4.9865 & \textcolor{blue}{\textbf{1.4201}}
        \\ 
        CDDFuse\cite{zhao2023cddfuse} & \textcolor{red}{\textbf{7.1203}} & \textcolor{blue}{\textbf{46.001}} & \textcolor{blue}{\textbf{13.1515}} & 2.1861 & 0.7682 & 0.5393 & \textcolor{blue}{\textbf{6.9245}} & 1.3411 &
        \textcolor{red}{\textbf{7.4366}} & \textcolor{red}{\textbf{54.669}} & \textcolor{blue}{\textbf{16.3633}} & 2.2996 & 0.6948 & 0.5242 & \textcolor{blue}{\textbf{8.2131}} & 1.3536 &
        \textcolor{blue}{\textbf{6.7012}} & \textcolor{red}{\textbf{43.3844}} & 11.5566 & \textcolor{red}{\textbf{3.4715}} & \textcolor{red}{\textbf{1.0509}} & \textcolor{blue}{\textbf{0.6929}} & 5.2811 & 1.3931\\ 
        SHIP\cite{zheng2024probing} & 6.9475 & 40.7861 & 12.9417 & \textcolor{red}{\textbf{2.8147}} & \textcolor{blue}{\textbf{0.7752}} & \textcolor{blue}{\textbf{0.5787}} & 6.8979 & 1.3114 & 7.1546 & 44.672 & 14.5852 & \textcolor{red}{\textbf{2.8331}} & \textcolor{blue}{\textbf{0.7072}} & 0.5763 & 7.7069 & 1.3425 & 6.4307 & 41.1334 & \textcolor{red}{\textbf{11.8149}} & 2.861 & 0.9075 & 0.6582 & \textcolor{red}{\textbf{5.5415}} & 1.3101 \\
        SMFNet (Ours) & \textcolor{blue}{\textbf{7.1106}} & 43.9525 & \textcolor{red}{\textbf{13.6222}} & 2.1574 & \textcolor{red}{\textbf{0.7846}} & \textcolor{red}{\textbf{0.584}} & \textcolor{red}{\textbf{7.2613}} & 1.3233 
        & \textcolor{blue}{\textbf{7.3909}} & 51.7062 & \textcolor{red}{\textbf{16.4399}} & 2.3024 & \textcolor{red}{\textbf{0.7328}} & \textcolor{blue}{\textbf{0.5831}} & \textcolor{red}{\textbf{8.6128}} & 1.3300
        & \textcolor{red}{\textbf{6.7021}} & 42.9562 & \textcolor{blue}{\textbf{11.6976}} & 3.1131 & \textcolor{blue}{\textbf{1.0448}} & \textcolor{red}{\textbf{0.7097}} & \textcolor{blue}{\textbf{5.4765}} & 1.3905
        \\ \bottomrule

\end{tabular}
}
\end{table*}

\subsection{Infrared-Visible Object Detection ($\mathbf{RQ2}$)}
\subsubsection{Setup} To rigorously compare the efficacy of fusion algorithms on object detection tasks, the data set and the detection method are fixed. The $M^{3}FD$ dataset \cite{liu2022target} is utilized to assess the impact of the proposed fusion model on object detection tasks. This dataset comprises 4200 pairs of infrared and visible images, which are divided into training, validation, and testing sets at a ratio of 8:1:1, respectively. The dataset encompasses six categories: people, cars, buses, motorbikes, trucks, and lamps. YOLOV5x\footnote{YOLOV5: \url{https://github.com/ultralytics/yolov5}} is employed as the object detector. The training process spans 100 epochs, with a batch size of 8 and the SGD optimizer set with an initial learning rate of ${1e}^{-2}$. The primary evaluation metric employed is the Mean Average Precision (MAP) at an Intersection over Union (IoU) threshold of 0.5 (MAP@0.5). This metric quantifies the model's ability to detect objects within images accurately.

\begin{table}[h!]
\centering
\caption{Detection comparisons of our method with eleven state-of-the-art methods on $M^{3}FD$ \cite{liu2022target} dataset. \textcolor{red}{\textbf{Red}} indicates the best results, and \textcolor{blue}{\textbf{blue}} shows the second-best values.}
\label{det_table}
\resizebox{\linewidth}{!}{
\begin{tabular}{ l|  c  c  c  c  c  c | c }
    \toprule
        \multicolumn{1}{c|}{\multirow{2}{*}{\textbf{Method}}} & \multicolumn{6}{c|}{\textbf{AP@0.5}} & \multirow{2}{*}{\textbf{mAP@0.5}} \\ 
        ~ & People & Car & Bus & Lamp & Motor & Truck & ~ \\ 
        \midrule
        DeFusion\cite{liang2022fusion} & 0.644 & 0.888 & 0.823 & 0.799 & 0.628 & 0.796 & 0.763 \\ 
        DenseFuse\cite{li2018densefuse} & 0.696 & 0.911 & 0.847 & 0.856 & 0.627 & 0.828 & 0.794 \\ 
        DIDFuse\cite{zhao2020didfuse} & 0.649 & 0.911 & 0.832 & 0.846 & 0.678 & 0.829 & 0.791 \\ 
        ReCoNet\cite{huang2022reconet} & 0.697 & 0.914 & 0.855 & \textcolor{blue}{\textbf{0.886}} & 0.7 & 0.842 & 0.816 \\ 
        MURF\cite{xu2023murf} & 0.641 & 0.872 & 0.806 & 0.809 & 0.645 & 0.759 & 0.755 \\ 
        TarDal\cite{liu2022target} & 0.647 & 0.878 & 0.82 & 0.779 & 0.639 & 0.844 & 0.768 \\ 
        U2Fusion\cite{xu2020u2fusion} & 0.499 & 0.681 & 0.482 & 0.136 & 0.318 & 0.498 & 0.436 \\ 
        PMGI\cite{zhang2020rethinking} & 0.606 & 0.866 & 0.786 & 0.735 & 0.484 & 0.751 & 0.705 \\ 
        SwinFusion\cite{ma2022swinfusion} & 0.69 & \textcolor{blue}{\textbf{0.919}} & 0.85 & 0.868 & 0.683 & 0.83 & 0.807 \\ 
        CDDFuse\cite{zhao2023cddfuse} & \textcolor{blue}{\textbf{0.714}} & \textcolor{blue}{\textbf{0.919}} & \textcolor{blue}{\textbf{0.86}} & 0.877 & 0.71 & 0.835 & 0.819 \\ 
        SHIP\cite{zheng2024probing} & 0.683 & 0.914 & 0.859 & 0.883 & \textcolor{blue}{\textbf{0.727}} & \textcolor{red}{\textbf{0.862}} & \textcolor{blue}{\textbf{0.821}}\\
        SMFNet & \textcolor{red}{\textbf{0.723}} & \textcolor{red}{\textbf{0.925}} & \textcolor{red}{\textbf{0.862}} & \textcolor{red}{\textbf{0.903}} & \textcolor{red}{\textbf{0.738}} & \textcolor{blue}{\textbf{0.856}} & \textcolor{red}{\textbf{0.835}} \\ 
        \bottomrule
\end{tabular}
}
\centering
\end{table}
\begin{figure*}[h]
    \centering
    \includegraphics[width=\textwidth]{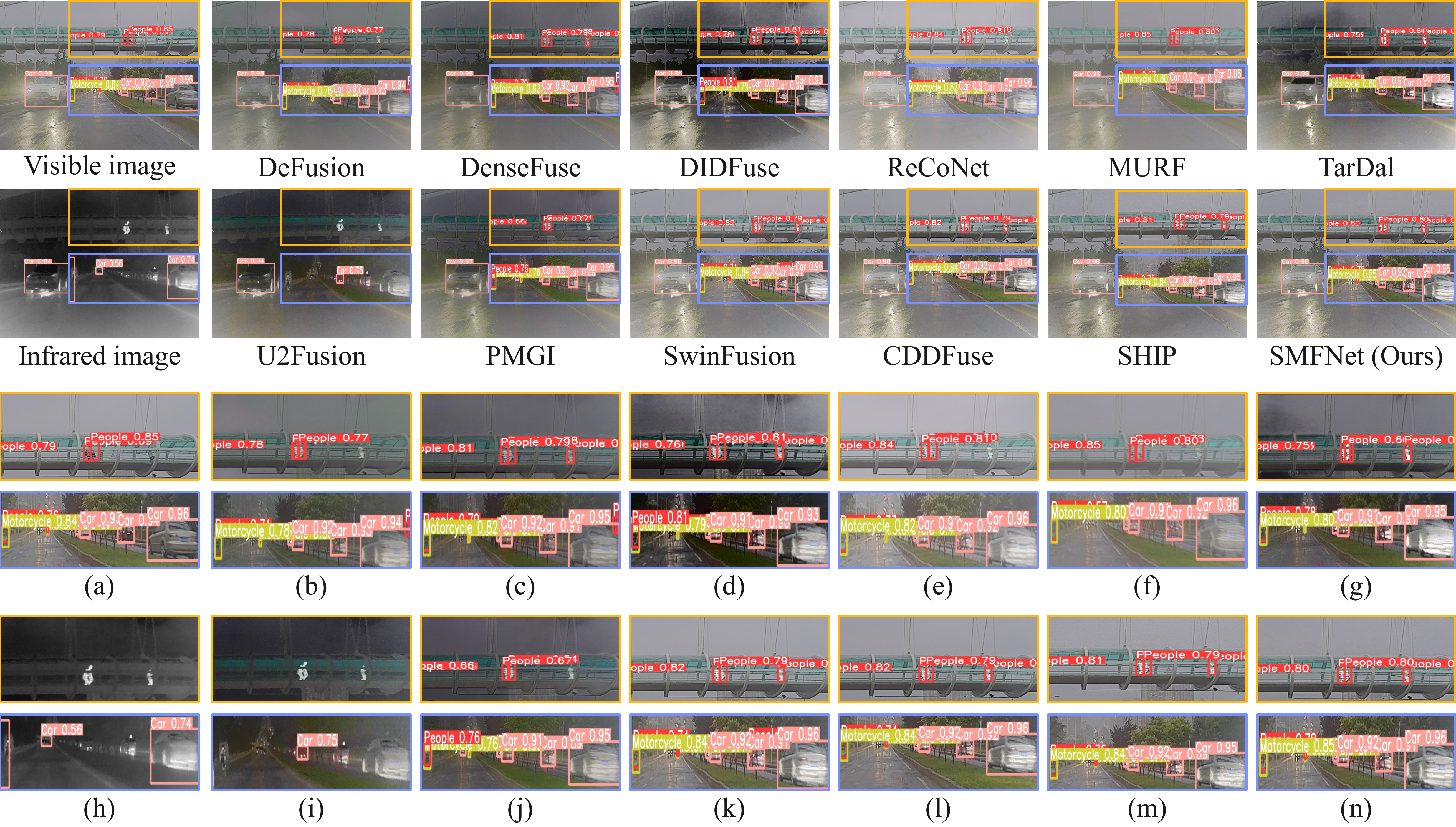}
    \caption{Qualitative detection results on ${M}^{3}FD$ dataset \cite{liu2022target}.  Magnified areas from (a) to (n) show the detection results of people walking on the pedestrian bridge, the motorbike and cars on the road, corresponding to the visible, infrared, and fused images generated by the state-of-the-art methods.}
    \label{M3FD}
\end{figure*}

\subsubsection{Compared with SOTA methods} Tab. \ref{det_table} and Fig. \ref{M3FD} display quantitative and qualitative results concerning object detection on $M^{3}FD$ dataset. SHIP\cite{zheng2024probing} has the best detection performance on the truck. ReCoNet\cite{huang2022reconet}, SwinFusion\cite{ma2022swinfusion} and CDDFuse\cite{zhao2023cddfuse} also show their strengths in detecting major objects in real street scenarios, especially for the cars and buses. Benefiting from the robust multi-guided architecture, the proposed model has the best detection performance in five classes, which demonstrates that generated fused images with specific characteristics can not only highlight targets difficult to detect but also preserve fundamental texture information.

\subsection{Infrared-Visible Segmentation ($\mathbf{RQ2}$)}
\subsubsection{Setup} DeepLabv3+ \cite{chen2018encoder} is employed as the backbone to conduct semantic segmentation. MFNet dataset \cite{ha2017mfnet} is divided into training and test sets following the settings in \cite{liu2023multi}. The model is supervised by cross-entropy loss with SGD optimizer and initial learning rate of ${1e}^{-2}$. There are totally 300 epochs. The mean accuracy (mACC) and mean intersection-over-union (mIoU) are employed as the evaluation metrics for segmentation quality.

\begin{table*}
\centering
\caption{Segmentation comparisons of our method with nine state-of-the-art methods on $MFNet$\cite{ha2017mfnet} dataset. \textcolor{red}{\textbf{Red}} indicates the best results, and \textcolor{blue}{\textbf{blue}} shows the second-best values.}
\label{seg_table}
\resizebox{0.8\linewidth}{!}{
\begin{tabular}{ l|  c  c  c  c  c  c  c  c | c  c}
    \toprule
        \multicolumn{1}{c|}{\multirow{2}{*}{\textbf{Method}}} & \multicolumn{8}{c|}{\textbf{IoU}} & \multirow{2}{*}{\textbf{mAcc}} & \multirow{2}{*}{\textbf{mIoU}}\\ 

        ~ & Unlabel & Car & Person & Bike & Curve & Stop & Cone & Bump & ~ & ~ \\ \midrule
        DeFusion & 0.962  & 0.673  & 0.415  & 0.387  & 0.157  & 0.100  & 0.199  & 0.113  & 0.624  & 0.339  \\ 
        DenseFuse & 0.961  & 0.658  & 0.390  & 0.480  & 0.154  & 0.118  & 0.298  & 0.267  & 0.610  & 0.376  \\ 
        DIDFuse & 0.961  & 0.649  & \textcolor{blue}{\textbf{0.439}}  & 0.429  & 0.084  & 0.100  & 0.287  & 0.206  & 0.587  & 0.356  \\ 
        ReCoNet & 0.963  & 0.693  & 0.410  & \textcolor{blue}{\textbf{0.493}}  & \textcolor{blue}{\textbf{0.167}}  & 0.110  & 0.310  & 0.065  & 0.581  & 0.360  \\ 
        MURF & 0.957  & 0.592  & 0.285  & 0.379  & 0.054  & 0.086  & 0.284  & 0.008  & 0.567  & 0.294  \\ 
        TarDal & \textcolor{blue}{\textbf{0.964}}  & 0.697  & 0.414  & 0.448  & 0.102  & 0.091  & 0.267  & 0.093  & 0.628  & 0.347  \\ 
        U2Fusion & 0.944  & 0.317  & 0.348  & 0.048  & 0.005  & 0.023  & 0.116  & 0.122  & 0.569  & 0.214  \\ 
        PMGI & 0.957  & 0.573  & 0.296  & 0.294  & 0.061  & \textcolor{blue}{\textbf{0.124}}  & 0.201  & 0.229  & \textcolor{red}{\textbf{0.646}}  & 0.312  \\ 
        SwinFusion & \textcolor{red}{\textbf{0.965}}  & \textcolor{red}{\textbf{0.703}}  &\textcolor{red}{\textbf{0.440}}  & 0.483  & \textcolor{blue}{\textbf{0.167}}  & \textcolor{red}{\textbf{0.135}}  & \textcolor{blue}{\textbf{0.315}}  & \textcolor{blue}{\textbf{0.280}}  & 0.637  & \textcolor{blue}{\textbf{0.394}}  \\ 
        CDDFuse & \textcolor{red}{\textbf{0.965}}  & \textcolor{blue}{\textbf{0.702}}  & \textcolor{red}{\textbf{0.440}}  & 0.449  & 0.157  & 0.119  & 0.307  & \textcolor{red}{\textbf{0.338}}  & 0.629  & 0.390  \\ 
        SHIP & 0.960 & 0.621 & 0.342 & 0.447 & 0.130 & 0.094 & 0.303 & 0.154 & 0.628 & 0.342 \\
        SMFNet & \textcolor{red}{\textbf{0.965}}  & 0.700  & 0.427  & \textcolor{red}{\textbf{0.499}}  & \textcolor{red}{\textbf{0.187}}  & 0.122  & \textcolor{red}{\textbf{0.328}}  & 0.268  & \textcolor{blue}{\textbf{0.640}}  & \textcolor{red}{\textbf{0.397}} \\ \bottomrule
\end{tabular}
}
\centering
\end{table*}
\begin{figure*}[h]
    \centering
    \includegraphics[width=\textwidth]{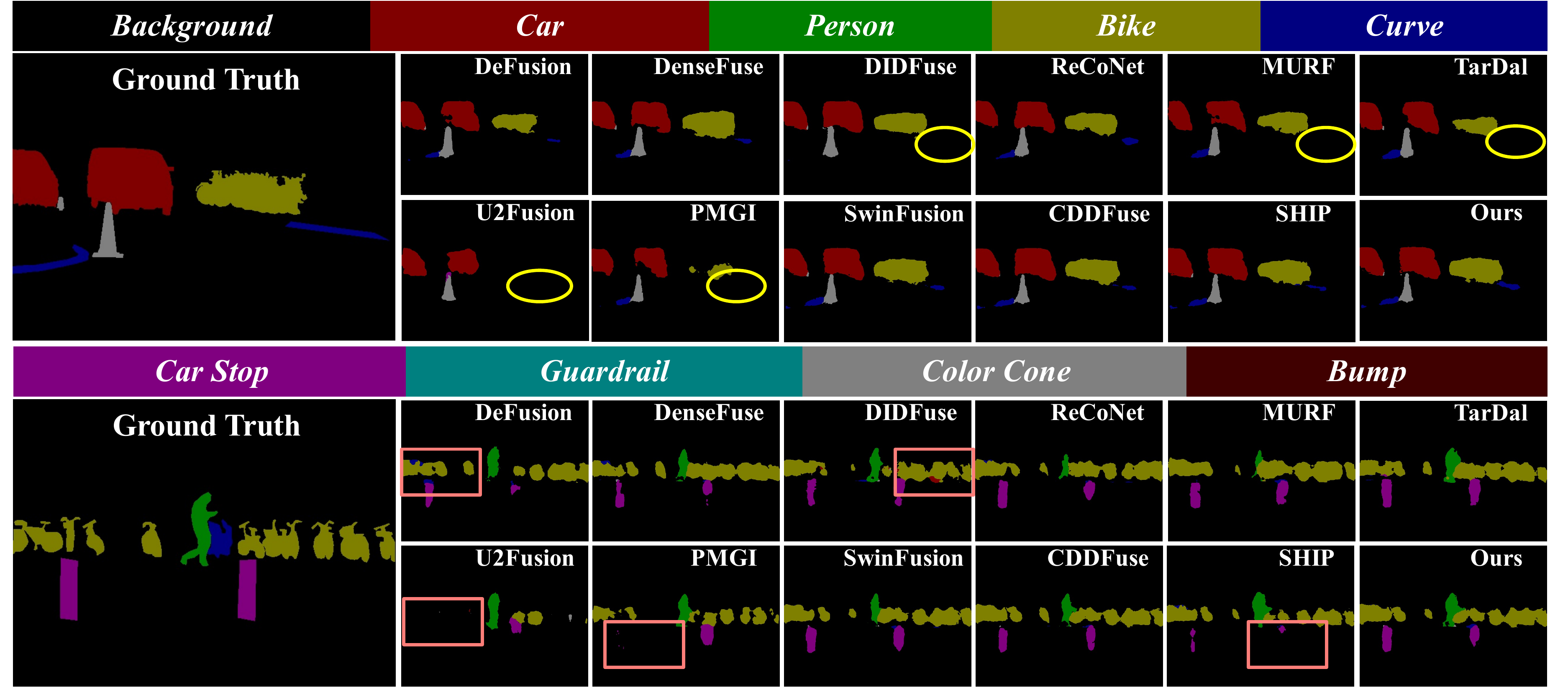}
    \caption{Visual comparison of semantic segmentation about fused images from nine fusion methods on the $MFNet$ \cite{ha2017mfnet} dataset. The yellow and pink regions show missing and error segmentation, respectively.}
    \label{segmentation}
\end{figure*}

\subsubsection{Compared with SOTA methods} The quantitative evaluation is shown in Tab. \ref{seg_table}. TarDal \cite{liu2022target}, SwinFusion \cite{ma2022swinfusion} and CDDFuse \cite{zhao2023cddfuse} enable to efficiently classify unlabelled pixel regions. Besides, PMGI \cite{zhang2020rethinking} has the highest segmentation accuracy. SwinFusion \cite{ma2022swinfusion} and CDDFuse \cite{zhao2023cddfuse} have the highest value of mIoU on recognizing the car and person classes. Our proposed model has shown competitive results, especially in the bike and cone regions. By observing Fig. \ref{segmentation}, Some methods, for example, DIDFuse \cite{zhao2020didfuse}, MURF \cite{xu2023murf}, TarDal \cite{liu2022target}, U2Fusion \cite{xu2020u2fusion} and PMGI \cite{zhang2020rethinking} have difficulties in distinguishing the curves and SHIP \cite{zheng2024probing} is hard to recognize car stops. SMFNet can better integrate the edge and contour information, and consequently makes semantic segmentation more accurate.

\begin{figure}[h]
    \centering
    \includegraphics[width=\linewidth]{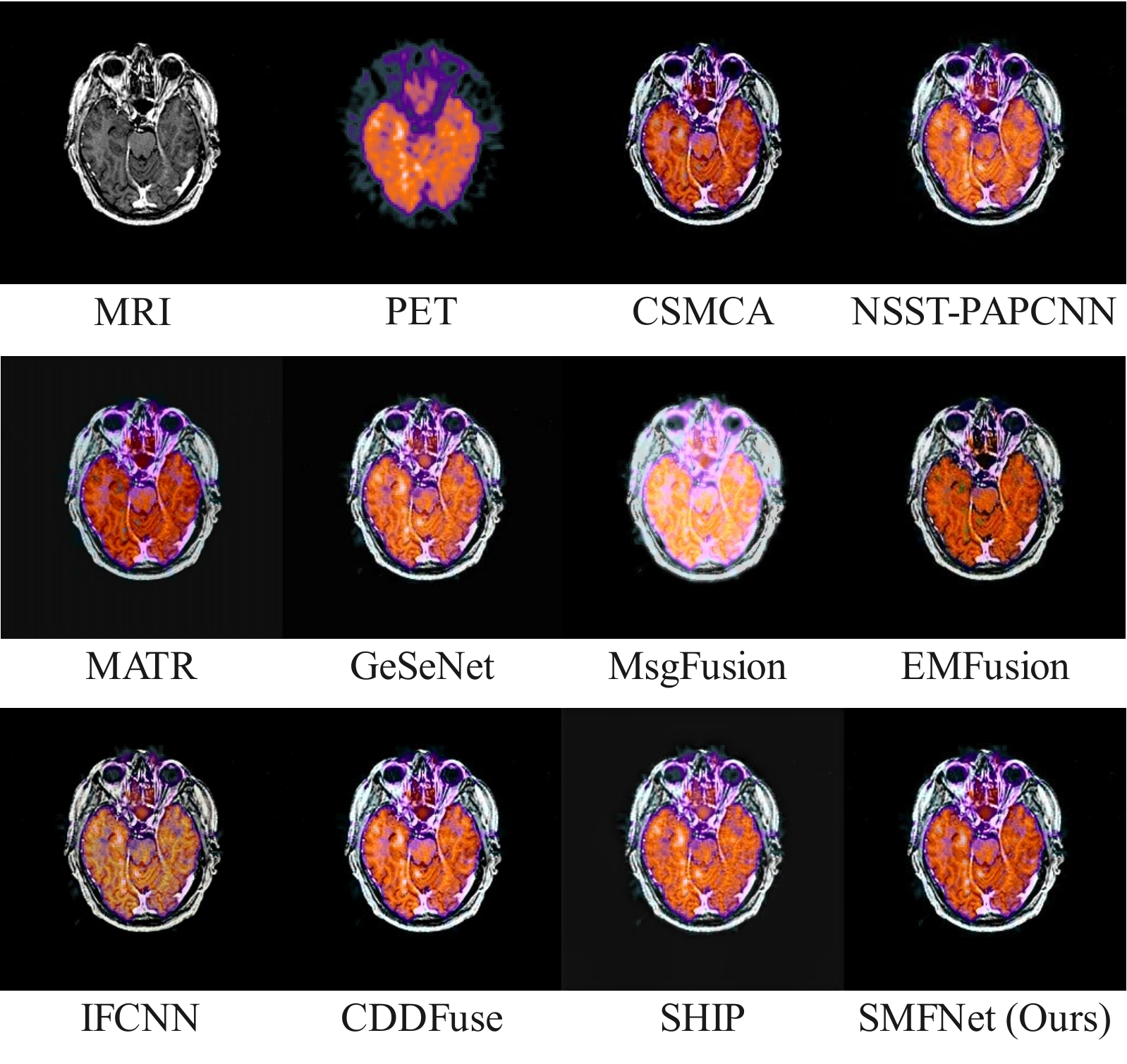}
    \caption{Fused images from the above eight fusion methods on $MIF$ dataset. CSMCA \cite{liu2019medical} and NSST-PAPCNN \cite{yin2018medical} are traditional algorithms, others are based on deep learning networks. SHIP \cite{zheng2024probing} is the result obtained without using the medical image dataset for training. Experiments show that our method retains soft tissue imaging features and contains rich color expressions.}
    \label{medical}
\end{figure}

\subsection{Extension to Medical Image Fusion ($\mathbf{RQ3}$)}
As an extension task to validate the generalizability of our methodology, we construct a dataset sourced from the Harvard Medical Website\footnote{\url{http://www.med.harvard.edu/AANLIB/home.html}}, including MRI-CT, MRI-PET and MRI-SPECT as training sets for the medical image fusion (MIF) task. 
There are 41 pairs of MRI-PET as testing sets.
Since the hyperparameters are configured identically to those of the VIF task, there is no need for a validation set. The compared methods are CSMCA \cite{liu2019medical}, NSST-PAPCNN \cite{yin2018medical}, MATR \cite{tang2022matr}, GeSeNet \cite{li2023gesenet}, MsgFusion \cite{wen2023msgfusion},
EMFusion \cite{xu2021emfusion} and IFCNN \cite{zhang2020ifcnn}. The fused medical images and objective evaluation results are listed in Fig. \ref{medical} and Tab. \ref{tab1}. The fused medical images compared in Fig. \ref{medical} demonstrate the superiority of the proposed model in persevering the structural information from MRI and the pseudo-color from PET. The objective evaluation results in Tab. \ref{tab1} further confirm that the proposed model can generate satisfactory fused images.
\begin{table}
\begin{center}
\caption{Extensive MRI-PET fusion experiment on Harvard Medical Dataset. ${\dagger}$ represents the results after training on MIF datasets. \textcolor{red}{\textbf{Red}} indicates the best results, and \textcolor{blue}{\textbf{blue}} shows the second-best values.}
\label{tab1}
\resizebox{\linewidth}{!}{
\begin{tabular}{ l | c  c  c  c  c  c  c  c }
\toprule
\multicolumn{1}{c|}{\multirow{2}{*}{\textbf{Method}}} & \multicolumn{6}{c}{\textbf{Medical Image Fusion Dataset}}\\
~ & MI &  SF & VIF & Qabf &  AG & SSIM \\ \midrule
        U2Fusion\cite{xu2020u2fusion} & 1.3879 & 12.2345 & 0.364 & 0.1933 & 3.8837 & 0.2086 \\ 
        TarDAL\cite{liu2022target} & 1.6094 & 18.1134 & 0.4268 & 0.3641 & 5.0768 & 0.2124 \\ 
        ReCoNet\cite{huang2022reconet} & 1.4408 & 11.1291 & 0.3717 & 0.1777 & 4.5917 & 0.2236 \\ 
        DeFusion\cite{liang2022fusion} & 1.5514 & 20.322 & 0.4439 & 0.4537 & 5.8414 & \textcolor{blue}{\textbf{1.5674}} \\ 
        CDDFuse\cite{zhao2023cddfuse} & \textcolor{blue}{\textbf{1.6226}} & \textcolor{blue}{\textbf{28.5224}} & \textcolor{blue}{\textbf{0.506}} & 0.606 & 7.7348 & \textcolor{red}{\textbf{1.7153}} \\
        SHIP\cite{zheng2024probing}  & 1.5968 & 28.0829 & 0.4754 & \textcolor{blue}{\textbf{0.632}} & \textcolor{blue}{\textbf{8.5999}} & 0.3374 \\
        SMFNet (Ours) & \textcolor{red}{\textbf{1.7284}} & \textcolor{red}{\textbf{32.6262}} & \textcolor{red}{\textbf{0.5881}} & \textcolor{red}{\textbf{0.7179}} & \textcolor{red}{\textbf{9.2834}} & 0.3267 \\ \midrule
        CSMCA$^{\dagger}$ \cite{liu2019medical} & 1.603 & \textcolor{blue}{\textbf{33.9865}} & 0.5604 & 0.6939 & 8.95 & 1.5506 \\ 
        NSST-PAPCNN$^{\dagger}$ \cite{yin2018medical} & 1.6361 & 31.4735 & 0.5407 & 0.6594 & 8.6443 & 1.4757 \\ 
        MATR$^{\dagger}$ \cite{tang2022matr} & 1.7516 & 25.6819 & \textcolor{red}{\textbf{0.689}} & 0.7202 & 7.3149 & 0.2793 \\ 
        GeSeNet$^{\dagger}$ \cite{li2023gesenet} & 1.6475 & 30.5983 & 0.5187 & 0.6844 & 8.8408 & 0.3049 \\ 
        MsgFusion$^{\dagger}$ \cite{wen2023msgfusion} & 1.6138 & 21.5543 & 0.3802 & 0.3814 & 6.0622 & 1.4825 \\ 
        EMFusion$^{\dagger}$ \cite{xu2021emfusion} & 1.644 & 30.4133 & 0.559 & 0.6879 & 8.4637 & 1.5531 \\ 
        IFCNN$^{\dagger}$ \cite{zhang2020ifcnn} & 1.5663 & 32.7834 & 0.5292 & 0.6874 & 8.9715 & 1.6058 \\ 
        CDDFuse$^{\dagger}$ \cite{zhao2023cddfuse} & \textcolor{blue}{\textbf{1.7681}} & \textcolor{red}{\textbf{34.9976}} & 0.6638 & \textcolor{blue}{\textbf{0.7327}} & \textcolor{red}{\textbf{9.674}} & \textcolor{red}{\textbf{1.7132}} \\ 
        SMFNet (Ours)$^{\dagger}$ & \textcolor{red}{\textbf{1.8031}} & 32.3699 & \textcolor{blue}{\textbf{0.6739}} & \textcolor{red}{\textbf{0.7366}} & \textcolor{blue}{\textbf{9.0584}} & \textcolor{blue}{\textbf{1.6155}} \\ \bottomrule

\end{tabular}
}
\end{center}
\end{table}

\begin{table*}[!h]
\centering
    \caption{Model complexity and running time on $MSRS$ dataset. }
    \label{model complexity and running time}
    \resizebox{\linewidth}{!}{
    \begin{tabular}{l | c c c c c c  c c c c c c}
    \toprule
       Metrics & DeFusion & DenseFuse & DIDFuse & ReCoNet & MURF & TarDal & U2Fusion & PMGI & SwinFusion & CDDFuse & SHIP & SMFNet \\ 
    \midrule
        Flops (G) & 81 & 27 & 115 & 7 & 185 & 91 & \textcolor{blue}{\textbf{1e-3}} & \textcolor{red}{\textbf{1e-6}} & 308 & 548 & 157 & 1462 \\ 
        Params (K) & 8043 & 74 & 373 & \textcolor{red}{\textbf{8}} & 116 & 297 & 659 & \textcolor{blue}{\textbf{42}} & 927 & 1783 & 526 & 4504 \\ 
        Time (s) & 0.05 & \textcolor{red}{\textbf{0.01}} & \textcolor{red}{\textbf{0.01}} & 0.03 & 0.7 & \textcolor{blue}{\textbf{0.02}} & 0.64 & 0.1 & 1.04 & 0.16 & 0.1 & 0.32 \\ 
    \bottomrule
\end{tabular}
}
\end{table*}

\subsection{Ablation Study ($\mathbf{RQ4}, \mathbf{RQ5}$ and $\mathbf{RQ6}$)}
In this subsection, we mainly validate the effectiveness of each module, feature decomposition and aggregation, and the reason for the two training stages. EN, SD, VIF and SSIM are employed as the objective metrics. Experiment results are shown in Tab. \ref{tab:freq}. 

\begin{figure}[h]
    \centering
    \includegraphics[width=\linewidth]{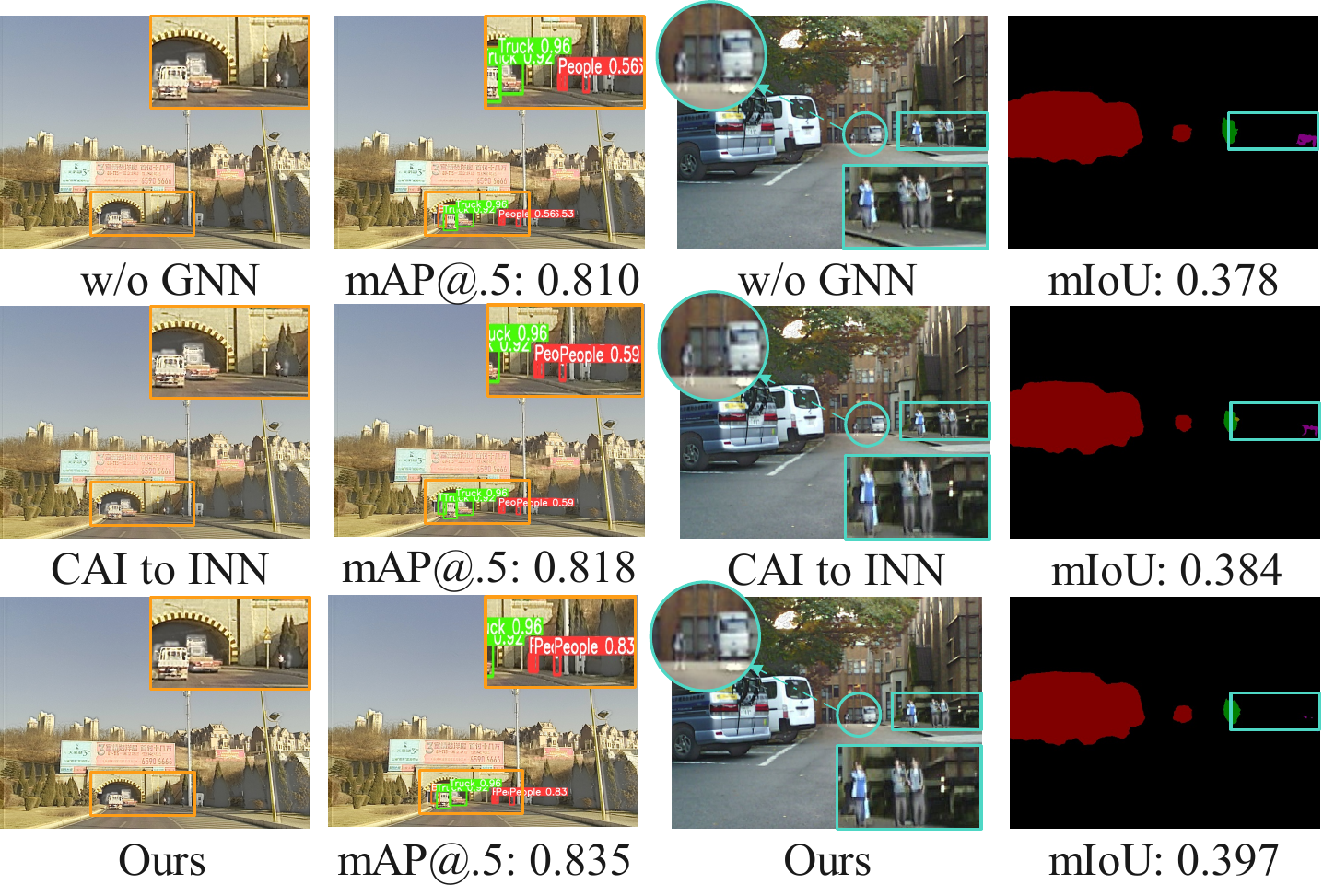}
    \caption{Ablation results of modules detection and segmentation on two datasets, $M^{3}FD$ (left) and $MFNet$ (right). The values of mIoU and mAP@.5 are listed as well.}
    \label{ablation}
\end{figure}

\subsubsection{Analysis on Modules}
To validate the proposed CAI's effectiveness, we replace it with the basic INN block (ID=AE1 in Tab. \ref{tab:freq}). The performance deteriorated as there was no cross-attention mechanism to enhance local semantic features. Note, different from the typical paradigm using cross attention to achieve cross-modal interaction, CAI is an implicit way to enhance high-frequency information. Additionally, we employed 4 stacked transformer blocks shown in Fig. \ref{fig:overflow} (d) without residual connections to capture low-frequency information in the encoder branch (ID=AE2). The results show that combining residual connections with CNNs improves long-range expression abilities. SMFNet without GNN (ID=AE3) leads to inadequate detailed feature extraction, which also affects performance across downstream tasks, as shown in Fig. \ref{ablation}. Besides, we changed the addition $\oplus$ operation at the end of the encoder into the concatenation 
$\raisebox{0.1em}{\begin{tikzpicture}[baseline=(C.base)]\draw (0,0) circle (0.12);\node (C) at (0,0) {\tiny\textbf{C}};\end{tikzpicture}}$
operation (ID=AE4). The $\raisebox{0.1em}{\begin{tikzpicture}[baseline=(C.base)]\draw (0,0) circle (0.12);\node (C) at (0,0) {\tiny\textbf{C}};\end{tikzpicture}}$ operation increases computational cost and adds complexity of handling high-dimensional features. Lastly, in order to verify the proper fusion rules, we swapped two fusion layers (ID=AE5) to determine their optimal positions for the fusion stage.

\begin{table}
\centering
\caption{Ablation experiment results in the test set of RoadScene \cite{xu2020u2fusion}. \textcolor{red}{\textbf{Red}} indicates the best results, and \textcolor{blue}{\textbf{blue}} shows the second-best values.}
\label{tab:freq}
\resizebox{\linewidth}{!}{
\begin{tabular}{c |l|  c  c  c  c | c }
    \toprule
      ID&  \multicolumn{1}{c|}{Configurations} & EN & SD & VIF & SSIM & \multicolumn{1}{c}{mAP@0.5} \\ \midrule
      AE1 & from CAI to INN block & 7.3257 & 49.5447 & 0.7139 & 1.3287 & 0.818 \\ 
      AE2 &BFE w/o residual connections & 7.3592 & 50.4884 & 0.7056 & 1.3023 & 0.82\\ 
      AE3 &w/o GNN & 7.3692 & 50.9436 & 0.7027 & 1.2896 & 0.810 \\
      AE4 & from addition to concatenation & \textcolor{blue}{\textbf{7.3773}} & 50.7761 & \textcolor{red}{\textbf{0.7893}} & 1.2982 & 0.821 \\
      AE5 & exchange detail and base fusion layer & 7.3584 & 50.8266 & 0.7247 & 1.3265 & 0.818 \\  \midrule
      AE6 &w/o $\mathcal{L}_{semantic}^I$ & 7.3579 & 50.7396 & 0.7297 & 1.325 & 0.829 \\ 
      AE7 & two-stage w/o $CC(\Phi_{G}^{V}, \Phi_{G}^{I})$ & 7.338 & 49.8241 & \textcolor{blue}{\textbf{0.7329}} & 1.33 & 0.825 \\ 
      AE8 & two-stage both with $CC(\Phi_{G}^{V}, \Phi_{G}^{I})$ & 7.3765 & \textcolor{blue}{\textbf{51.4596}} & 0.7248 & 1.3165 & 0.824 \\ 
      AE9 &w/o two-stage training & 7.365 & 50.4629 & 0.6906 & 1.3102 & 0.819 \\ \midrule
      AE10 & $\alpha_{1}, \alpha_{3}=1$ & 7.356 & 50.4861 & 0.7188 & \textcolor{blue}{\textbf{1.3303}} & \textcolor{blue}{\textbf{0.83}} \\ 
      AE11 & $\alpha_{1}, \alpha_{3}=5$ &  7.3788 & 50.8299 & 0.7126 & 1.3168 & 0.825\\ 
      AE12 & $\alpha_{1}, \alpha_{3}=8$ &  7.3732 & 51.3246 & 0.7265 & \textcolor{red}{\textbf{1.333}} & 0.829 \\ 
      AE13 & $\alpha_{1}, \alpha_{3}=10$ &  7.3791 & 50.9621 & 0.6999 & 1.2858 & 0.816 \\  \midrule
      AE14 &SMFNet & \textcolor{red}{\textbf{7.3909}} & \textcolor{red}{\textbf{51.7062}} & 0.7328 & 1.33 & \textcolor{red}{\textbf{0.835}} \\ \bottomrule
\end{tabular}
}
\centering
\end{table}

\begin{figure}[h]
    \centering
    \includegraphics[width=\linewidth]{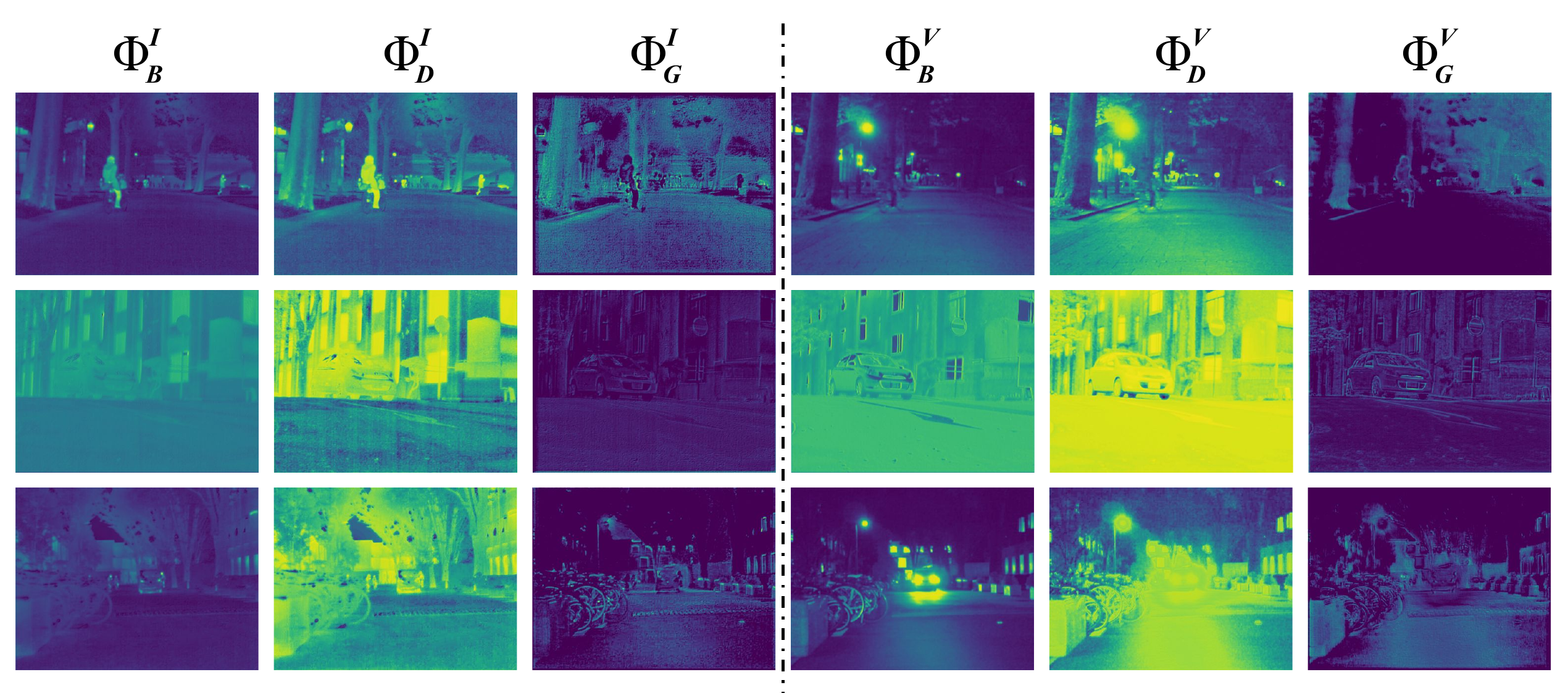}
    \caption{Feature decomposition results on $MSRS$ dataset. The three rows correspond to ``00706N'', ``01603D'' and ``01110N'' (from top to bottom).}
    \label{feature_map}
\end{figure}

\subsubsection{Features Decomposition and Aggregation}
In this part, we discuss the proposed semantic loss in training stage I (ID=AE6) and the correlation-driven GNN loss proposed in training stage II (ID=AE7 and AE8). During the experiment, we found that the absence of $CC(\Phi_{G}^{V}, \Phi_{G}^{I})$ to guide the fusion of GNN features in the second stage hinders the fusion performance. As a comparison, two training stages both with the guidance of correlation loss function, made the fused images reduce information preservation and have low contrast. We hypothesize that GNN does not need explicit correlation restriction when executing the image reconstruction task\cite{Li2025}. And in the fusion stage, it's necessary to increase correlations between nodes $\mathbf{C}$ and $\mathbf{D}$ (explained in Sec. \ref{sec3}) as detail features. Gram matrix brings salient targets and maintains abundant texture details. $CC(\Phi_{G}^{V}, \Phi_{G}^{I})$ in the second stage further enhances cross-modality interaction and balance between high-frequency and low-frequency information. From Fig. \ref{feature_map}, we provide the feature maps extracted from the three-branches encoder. Base features $\Phi_{B}^{I}$ and $\Phi_{B}^{V}$ contain gradual intensity changes with fewer sharp edges. Detail features $\Phi_{D}^{I}$ and $\Phi_{D}^{V}$ capture abrupt intensity changes and emphasize local contrast. $\Phi_{G}^{I}$ and $\Phi_{G}^{V}$ extracted from GR module highlight the edge and boundary of the objects and are also as the complementary of detail features. 

\subsubsection{Two Training Stages}
To validate the importance of two training stages, we directly train the encoder, decoder and fusion layers together (ID=AE9) with 120 epochs. The results are unsatisfactory. Pre-trained encoder and decoder show more powerful feature extraction abilities and promote training in the second stage to generate better performance on qualitative and quantitative results. To validate the weighting of decomposition loss, we set $\alpha_{1}, \alpha_{3}=1, 5, 8, 10$, respectively (from ID=AE10 to ID=AE13). The settings we chose $\alpha_{1}, \alpha_{3}=2$ is optimal. The $\mathcal{L}_{decomp}^I$ and $\mathcal{L}_{decomp}^{II}$ play key roles in guiding the network to obtain decomposition features for the reconstruction and fusion procedure. 

\begin{table}
\centering
\caption{Quantitative evaluation results on $MSRS$ dataset compared with big model based methods, where \textcolor{red}{\textbf{red}} indicates the best results, and \textcolor{blue}{\textbf{blue}} shows the second-best values.}
\label{big model}
\resizebox{\linewidth}{!}{
\begin{tabular}{ l | c  c  c  c  c  c  c  c | c  c  c  }
    \toprule
    Method & EN &  SD &  SF &  MI & VIF & Qabf &  AG & SSIM & GFlops & Params & Time\\ \midrule
    Text-IF\cite{yi2024text} & \textcolor{blue}{\textbf{6.7281}} & \textcolor{blue}{\textbf{44.5889}} & \textcolor{red}{\textbf{11.8784}} & \textcolor{blue}{\textbf{2.8202}} & \textcolor{blue}{\textbf{1.0311}} & \textcolor{blue}{\textbf{0.6747}} & \textcolor{blue}{\textbf{5.4684}} & \textcolor{red}{\textbf{1.3913}} & \textcolor{blue}{\textbf{1519}} & \textcolor{blue}{\textbf{89014}} & \textcolor{red}{\textbf{0.21}} \\ 
    Text-DiFuse\cite{zhang2025text} & \textcolor{red}{\textbf{7.1435}} & \textcolor{red}{\textbf{54.2258}} & 11.406 & 1.9698 & 0.7306 & 0.4406 & 5.212 & 0.9602 & 8604 & 119398 & 15 \\ 
    SMFNet & 6.7021 & 42.9562 & \textcolor{blue}{\textbf{11.6976}} & \textcolor{red}{\textbf{3.1131}} & \textcolor{red}{\textbf{1.0448}} & \textcolor{red}{\textbf{0.7097}} & \textcolor{red}{\textbf{5.4765}} & \textcolor{blue}{\textbf{1.3905}} & \textcolor{red}{\textbf{1462}} & \textcolor{red}{\textbf{4504}} & \textcolor{blue}{\textbf{0.32}} \\ 
         \bottomrule

\end{tabular}
}
\end{table}

\subsection{Discussion}
\subsubsection{Efficiency Comparison}
Tab. \ref{model complexity and running time} lists the computational costs and execution time of the twelve image fusion methods on the 361 image pairs of the MSRS test set with the size of $640 \times 480$. MURF, U2Fusion and PMGI run on TensorFlow platform with CPU. Other methods run on Pytorch with GPU acceleration. By numerical comparison, U2Fusion and PMGI have the minimal computational load. Additionally, The model of ReCoNet only has 8K parameters. DenseFuse and DIDFuse spend an average of 0.01 seconds generating per fused image. The proposed SMFNet spends less generating time than SwinFusion and has fewer parameters than DeFusion. In Tab. \ref{big model}, Text-IF \cite{yi2024text} and Text-DiFuse\cite{zhang2025text} have the largest model parameters and computational cost owing to the help of big model, e.g., CLIP\cite{yi2024text} and diffusion model\cite{VDMU}.                             

\subsubsection{More insights on our model}
SMFNet exhibits stable performance on infrared-visible image fusion task and the fused images can also be applied into downstream tasks. In this paper, we decomposed features into high-frequency and low-frequency features by distinct architecture (i.e., CNN, Transformer and GNN). The correlation coefficient constrains features of the same type from two modalities to enhance detail preservation and saliency information in a multi-guided manner. Recent works proposed an overlooked issue: the over-reliance on visual features in the realm of image fusion\cite{zhao2024image}. Some pioneers\cite{zhao2024image}, \cite{cheng2025textfusion}, \cite{yi2024text}, \cite{zhang2025text} extracted textual semantic information and guided the extraction and fusion procedures at the vision level. In Tab. \ref{big model}, the big model guided methods quantitative results on MSRS dataset are listed as a fair comparison. Notably, a complex network with high computational cost is challenging to be deployed on small-footprint devices, e.g., mobile phones. The utilization of text features and the computational overhead are our concerns to focus on for further improvements.
\section{CONCLUSION}
\label{sec5}
In this paper, a semantic-aware multi-guided network for infrared-visible image fusion was proposed. Graph Reasoning module models and reasons about high-level relations between two modalities and extract low-level detail features. Cross Attention and Invertible Block (CAI) solves information loss during the decomposing process to extract high-frequency features. Additionally, a base feature extraction module is primarily utilized to capture global features with long-range dependencies. Two novel loss functions are proposed as the restriction in two training stages. Experiments demonstrate the fusion effect of our proposed SMFNet and the accuracy of downstream pattern recognition tasks can also be improved. In further works, we will explore multi-task learning (MTL) including image reconstruction, fusion tasks and scene parsing, considering novel joint-optimization among these tasks soon.


 




\vfill

\end{document}